\documentclass{article}

\usepackage[main, final]{neurips_2026}

\usepackage[utf8]{inputenc}
\usepackage[T1]{fontenc}
\usepackage{hyperref}
\usepackage{url}
\usepackage{booktabs}
\usepackage{amsfonts}
\usepackage{amsmath}
\usepackage{amssymb}
\usepackage{mathtools}
\usepackage{amsthm}
\usepackage{nicefrac}
\usepackage{microtype}
\usepackage{xcolor}
\usepackage{graphicx}
\usepackage{subcaption}
\usepackage[capitalize,noabbrev]{cleveref}

\theoremstyle{plain}

\theoremstyle{definition}

\theoremstyle{remark}

\usepackage[disable,textsize=tiny]{todonotes}

\title{STEP: Learning STructured Embeddings for Progressive Time Series}

\author{%
  Lucas Thil \\
  LIX, École Polytechnique \\
  IRT SystemX \\
  Palaiseau, France \\
  \texttt{thil@lix.polytechnique.fr} \\
  \And
  Jesse Read \\
  LIX, École Polytechnique \\
  Palaiseau, France \\
  \And
  Rim Kaddah \\
  IRT SystemX \\
  Palaiseau, France \\
  \And
  Guillaume Doquet \\
  Safran Tech \\
  Châteaufort, France \\
}

\begin{document}

\addtocontents{toc}{\protect\setcounter{tocdepth}{-10}}

\maketitle

\begin{abstract}
We present a novel method for learning structured representations of progressive time series, that is, data capturing irreversible state transitions such as degradation or task completion. Our approach uses a self-supervised contrastive objective to learn a low-dimensional latent space in which each observation becomes a point on a manifold anchored between two fixed orthogonal prototype vectors, and a trajectory becomes a path across that manifold. From this structure we read a \emph{latent compass}, the polar coordinates $(\theta, r)$ of the latent vector, in which $\theta$ tracks the progression of the underlying state (e.g., from healthy to failed) and $r$ 
exposes residual, non-progression structure, without a supervised progression target.
We evaluate the approach on industrial degradation and robotic tasks, validating three key capabilities: (1) end-state prediction, (2) health indicators that track the degradation state, and (3) phase separation. Our method is competitive with or improves over black-box counterparts while exposing the progression through a fixed set of coordinates. A simple linear regressor on top of the learned coordinates is competitive with deep architectures, quantitative evidence that the underlying state is encoded in a geometrically accessible form. Code is available at \url{https://github.com/LucasStill/STEP}.
\end{abstract}

\section{Introduction}

Time series from critical systems such as aircraft engines, batteries, or robots often capture \emph{progressive} trajectories: a system evolves from an initial state (e.g., healthy) to a distinct end state (e.g., degraded, or task completed). In this paper, a \textbf{progressive time series} is a sequence of noisy, high-dimensional measurements $x_t$ of an unobserved state $s_t$, $t=1,\ldots,T$, that follows a non-regressive transition: as $t$ increases, $s_t$ moves away from its initial distribution and does not return to it under nominal operation. The measurements themselves may oscillate because of sensor noise or changing operating conditions.

The canonical application is prognostics and health management (PHM)~\citep{Lee_Wu_Zhao_Ghaffari_Liao_Siegel_2014}, where one estimates the Remaining Useful Life (RUL, the number of steps before a terminal event such as failure, $T-t$) or a State of Health (SOH) that evolves monotonically with $s_t$. Two difficulties arise. First, $s_t$ is never observed, and $x_t$ need not reflect its progressive nature directly. Second, the only training signal is time itself: in expectation the system degrades from $t$ to $t+1$, and the terminal state is reached at $T$.

Existing approaches face a trade-off. Hand-crafted health indicators (HIs) and Self-Organizing Maps (SOMs) are transparent but inflexible~\citep{tsui2015prognostics,bougacha2020review}. Deep models predict RUL or SOH accurately, but their latent spaces are hard to inspect, which matters in safety-critical settings, and they are trained against proxy targets such as capped RUL, which can be noisy or sparse in real fleets.

We seek a latent representation $z_t$ that (i) supports strong downstream performance, e.g., RUL prediction, and (ii) exposes progression through fixed-formula coordinates of $z_t$ rather than through post-hoc explanations. We say \emph{structured} rather than \emph{interpretable} on purpose: we show that these coordinates are low-dimensional, predictive and linearly accessible, not that they are physically meaningful or mechanistically faithful health measurements.

To this end we propose \textbf{STEP} (\textbf{ST}ructured \textbf{E}mbeddings for \textbf{P}rogressive Time Series). STEP learns a low-dimensional latent space in which trajectories run between two fixed orthogonal prototype vectors $z_\text{init}, z_\text{end}$. Two readouts follow directly from $z_t$: the polar coordinates of $z_t$ in the prototype plane, a \emph{latent compass} whose angle $\theta_t$ tracks progression and whose radius $r_t$ exposes residual structure; and a full-dimensional Mahalanobis distance from the healthy state. Both are computable at inference without oracle knowledge of $T$. Our core contributions are:
\begin{enumerate}
    \item A new paradigm for learning representations of progressive time series, framing state evolution as geometric paths in a contrastively learned latent space without a supervised progression target.
    \item A readout tool (that we call a `latent compass', essentially high-level features) that exposes the progression of the underlying state through a fixed set of coordinates.
    \item Demonstration of competitive performance on downstream tasks (RUL regression and health indicators) across diverse degradation benchmarks. The learned coordinates enable simple linear models to match or outperform complex architectures, evidence that the state is encoded in a geometrically accessible form.
\end{enumerate}

\begin{figure}[t]
    \centering
    \includegraphics[width=\textwidth]{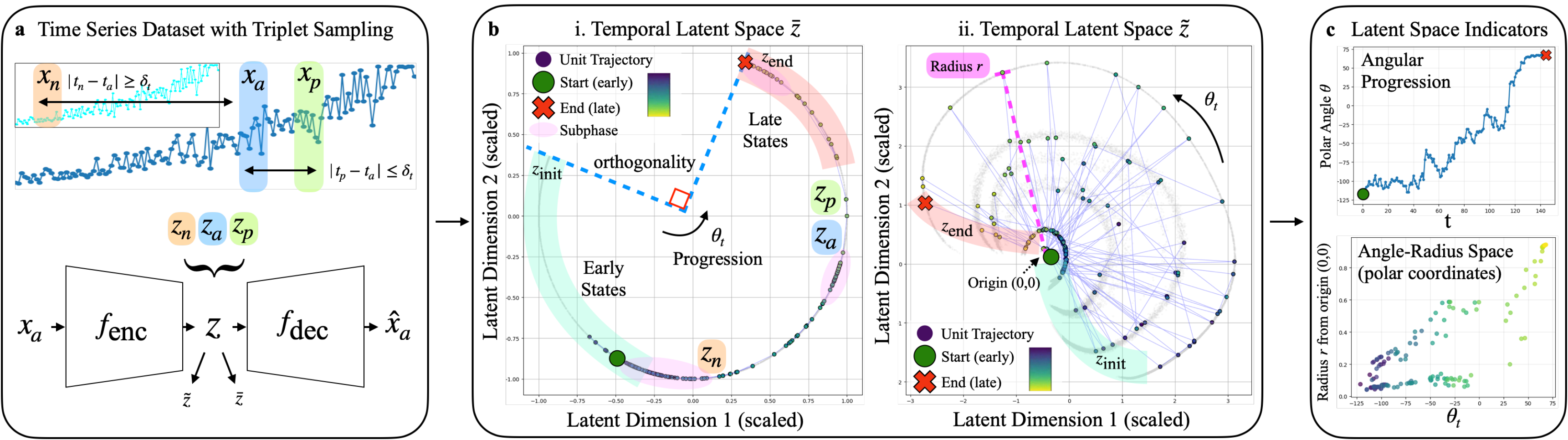}
    \caption{\textbf{Approach overview.} \textbf{a) Model learning:} an autoencoder is trained with triplets in which the anchor $x_a$ and positive $x_p$ are temporally close ($|t_p-t_a|\le\delta_t$) and the negative $x_n$ is temporally distant or from a different unit. \textbf{b) Latent space.} (i) The standardized view $\tilde{z}$ shows a progression between the fixed orthogonal prototypes $z_\text{init}$ and $z_\text{end}$, with sub-phase clusters along the manifold. (ii) The unit-normalized view $\bar{z}$ projects trajectories onto the unit circle. \textbf{c) Latent compass:} the angle $\theta_t$ and radius $r_t$ are deterministic functions of $z_t$ and serve as features for downstream tasks (RUL regression, phase clustering).}
    \label{fig:approach_overview}
\end{figure}

The paper is organized by first covering related work (Sec.~\ref{sec:related_works}), presenting our method (Sec.~\ref{sec:method}), detailing experimental protocols (Sec.~\ref{sec:experiments}), and discussing results (Sec.~\ref{sec:discussion}).

\section{Related Work}\label{sec:related_works}

\paragraph{Supervised RUL prediction.} A dominant strand of PHM frames prognostics as supervised RUL regression with CNN/LSTM hybrids~\citep{Zheng_Ristovski_Farahat_Gupta_2017,Peng_Chen_Chen_Tang_Li_Gui_2021,Li_Zhao_Zhang_Zio_2020}, uncertainty-aware variants~\citep{De_Pater_Mitici_2022,Costa_Sanchez_2022}, or autoencoder and SSL feature extraction~\citep{Pillai_Vadakkepat_2021,Listou_Ellefsen_Bjorlykhaug_AEsoy_Ushakov_Zhang_2019}. These models reach strong benchmark accuracy but return a scalar RUL without exposing the degradation progression. Threshold-based and parametric methods are transparent but less adaptable~\citep{kim2017prognostics}.

\paragraph{Survival analysis and time-to-event modeling.} RUL estimation is a time-to-event problem. Survival models describe the hazard of the terminal event, from proportional hazards~\citep{cox1972regression} to deep survival models~\citep{katzman2018deepsurv,lee2018deephit}, and naturally exploit right-censored units and competing risks. They output an event-time distribution given covariates; STEP instead learns a representation whose geometry orders states along the trajectory, and uses complete run-to-failure trajectories for training (Sec.~\ref{sec:conclusion}). 

\paragraph{Health indicators (HIs).} A parallel line builds scalar HIs that correlate monotonically with degradation~\citep{Rombach_Michau_Burzle_Koller_Fink_2024,thil2025glide,leyliabadi2026machinelearningframeworkturbofan}, often from reconstruction error or novelty scores, and evaluates them with monotonicity, trendability and prognosability~\citep{coble2010merging}. Many target end-of-life regimes~\citep{Zhao_Bin_Liang_Wang_Lu_2017,fu2025degradation} rather than the whole trajectory, and rely on predefined modes or labels that are often noisy or unavailable~\citep{Fink_Wang_Svensen_Dersin_Lee_Ducoffe_2020}. Closest in spirit, hierarchical controlled differential models separate slow degradation from fast operating fluctuations by conditioning on control inputs~\citep{Zhao_Fink_2027}.

\paragraph{Contrastive and topology-based representation learning.} Contrastive methods learn time-series representations from positive/negative pairs~\citep{Zhang_Zhao_Tsiligkaridis_Zitnik_2022,eldele2021time,sermanet2018tcn,baydogan2016time}, with refinements such as combined losses~\citep{Cai_Zhang_Liu_2025}, long-range dependencies~\citep{Park_Gwak_Choo_Choi_2024}, soft pair assignments~\citep{Liu_Chen_2023} and SoftCLT's soft temporal contrast~\citep{Lee_Park_Lee_2024}. They do not target a globally ordered progression geometry. SOMs~\citep{Come} and GTM~\citep{Bishop_Svensen_Williams_1998} are transparent but assume a fixed discrete grid; learned topologies~\citep{Mladenov_Martinez_Garcia_Zhang_Khan_Patankar_2023} and triplet-based stage representations~\citep{schroff2015facenet,Fan_Li_Guo_Yu_Luo_Gao_2024} are closest to our setting.

\paragraph{Progressive representations beyond PHM.} Task progression also provides long-horizon signals in robotics~\citep{Chen_Yu_Schwager_Abbeel_Shentu_Wu_2025,ma2023liv}, and recent latent-planning work argues for low-dimensional bottlenecks that straighten progressive trajectories~\citep{wang2026temporalstraightening}. Hyperspherical, cosine-based metric learning~\citep{li2017deep,kumar2016learning} provides the geometric tools we build on.

\section{STEP (STructured Embeddings for Progressive Time Series)}\label{sec:method}

The central idea behind our approach is to make progression readable directly from the geometry of the latent space. Each observation $x_t$ is encoded into a point $z_t$ on a learned manifold (latent space), and thus an entire trajectory becomes a path across that manifold, and two polar coordinates of $z_t$, namely the angle $\theta_t$ and the radius $r_t$, together form a \emph{latent compass}: $\theta_t$ measures how far along the progression a point is (e.g., from healthy to failed), while $r_t$ captures residual structure beyond the progression (e.g., the concentric operating-condition rings of Fig.~\ref{fig:latent_his_results}). Reading the compass thus tracks progression through a fixed, low-dimensional set of coordinates.

We learn this by combining three forces in a composite objective.
A contrastive cosine triplet loss orders observations by state similarity, producing a coherent, ordered manifold such that the derived coordinates allow localization of the observed states.
A prototype loss anchors the empirical start and end of every trajectory to a fixed orthogonal basis $\{z_\text{init}, z_\text{end}\}$, fixing a consistent frame across runs and datasets. A reconstruction term acts as a mild regularizer, preventing latent collapse. The full objective is

\begin{equation}
\label{eq:composite_loss}
\mathcal{L}_\text{total} =  \lambda_{1}\,\mathcal{L}_\text{recon} + \lambda_{2}\,\mathcal{L}_\text{trip} + \lambda_{3}\,\mathcal{L}_\text{proto}.
\end{equation}

with weights $\lambda_1,\lambda_2,\lambda_3$. Together, these terms make progression readable from a few fixed coordinates.
The next subsections explain each component, and empirical results follow in Sec.~\ref{sec:experiments} across multiple use cases.

\subsection{Setting and Goal}\label{sec:method_goal}

We observe $U$ independent progressive time series $\{(x^{(u)}_1,\ldots,x^{(u)}_{T_u})\}_{u=1}^U$ with $x^{(u)}_t \in \mathbb{R}^d$. Each $x_t$ is a noisy observation of an unobserved state $s_t$ that moves non-regressively from $s_1$ (e.g., healthy) to $s_T$ (e.g., failed or task completed); a Markov-chain formalization is given in Appendix~\ref{appx:notation}. There is no ground-truth label for $s_t$: at training time only the start and end of each complete trajectory are observable surrogates for the progression boundaries. We want an encoder $z_t = f_\text{enc}(x_t)$ whose geometry tracks $s_t$ rather than $t$: two observations in similar states should be close in $z$-space regardless of how much time each unit has accumulated. Sec.~\ref{sec:discussion_theta_vs_t} tests this on held-out units.

\subsection{A Contrastive Ordering Objective}\label{sec:method_triplet}

We use a triplet loss with cosine similarity. For a triplet $(x_a, x_p, x_n)$ with latent codes $(z_a, z_p, z_n)$,
\begin{equation}
\mathcal{L}_\text{trip} = \frac{1}{N} \sum_{i=1}^N \max\!\bigl(0,\ \cos(z_a, z_n) - \cos(z_a, z_p) + m\bigr),
\label{eq:triplet_loss}
\end{equation}
where $m$ is the margin and $\cos(z_a, z_p) = \tfrac{z_a \cdot z_p}{\|z_a\|\|z_p\|}$. Since $\cos(z_a, z_p) = \cos(\theta_{ap})$, Eq.~\eqref{eq:triplet_loss} encourages $\theta_{ap} < \theta_{an}$, ordering embeddings angularly along the trajectory.

We construct triplets using temporal proximity within a trajectory as a \emph{sampling proxy} for state similarity. The positive $x_p$ is drawn from the same unit as the anchor $x_a$, with $0 < |t_p - t_a| \le \delta_t$; the negative $x_n$ is drawn either from the same unit with $|t_n - t_a| \ge \delta_t$ or from a different unit entirely. Temporal proximity is a sampling heuristic: Sec.~\ref{sec:discussion_theta_vs_t} shows that the resulting geometry tracks state rather than collapsing to elapsed time. In practice, negatives are drawn from a different coarse remaining-life band of the complete training trajectories (${>}125$, $75$--$125$, $30$--$75$, ${\le}30$ cycles; Appendix~\ref{appx:training}). STEP thus uses no remaining-life regression target, but its pairing does use this coarse end-of-life information.

\paragraph{Per-observation encoder.} $f_\text{enc}$ processes each window $x_t$ independently; time enters only through sampling. This is the modular split of most SSL methods (SimCLR~\citep{chen2020simple}, BYOL~\citep{grill2020bootstrap}, SoftCLT~\citep{Lee_Park_Lee_2024}, LeJEPA~\citep{Balestriero_LeCun_2025}): the loss shapes the geometry of single-input representations and any sequence modeling is left to the downstream head. The same frozen encoder therefore serves several tasks, and every readout is a deterministic function of a single $z_t$.

We add a mean squared error term $\mathcal{L}_\text{recon} = \frac{1}{N}\sum_{i}\|x_i-f_\text{dec}(f_\text{enc}(x_i))\|^2_2$ with a deliberately small MLP decoder, so that a high-capacity decoder cannot bypass the latent structure. Alone, this term yields a weakly ordered latent (Sec.~\ref{sec:hi_quality}).

\subsection{A Fixed Reference Frame}\label{sec:method_proto}

The triplet objective orders the manifold but leaves it free to rotate, reflect and rescale across runs, so the same observation can land at different coordinates from one training to the next, and $\theta$ and $r$ cannot be compared across runs or datasets. We therefore fix two standard basis vectors as anchors, $z_\text{init} = \mathbf{e}_1$ and $z_\text{end} = \mathbf{e}_2$, and softly pull the first $k$ and last $k$ steps of each training trajectory, $X_\text{init}$ and $X_\text{end}$, toward them:
\begin{equation}
\mathcal{L}_\text{proto} = \mathbb{E}_{x \in X_\text{init}}\!\left[\|f_\text{enc}(x) - z_\text{init}\|^2\right] + \mathbb{E}_{x \in X_\text{end}}\!\left[\|f_\text{enc}(x) - z_\text{end}\|^2\right].
\end{equation}
The latent space thus contains an initial-state region (around $\mathbf{e}_1$) and an absorbing terminal region (around $\mathbf{e}_2$), with trajectories navigating between them.
Intermediate states are never conditioned on. The pull is a soft expectation: observations cluster \emph{near} the anchors, leaving slack for unit-level variability such as different initial wear (the spread of the healthy zone in Fig.~\ref{fig:approach_overview}), and the radial direction is left free. Anchoring is not always needed for accuracy but matters when trajectories span heterogeneous regimes: removing it ($\lambda_3{=}0$) degrades FD004 from 12.10 to 19.01 RMSE but slightly improves the homogeneous FD003 (13.35 to 11.59; single runs, Appendix~\ref{appx:proto_ablation}).

Together, the contrastive ordering and the prototype-anchored basis induce a \emph{state-based topology} over the dataset, in which each observation projects to a single \emph{step} along the trajectory map, from start to end.

\subsection{Readouts}\label{sec:readouts}

We read two standard views of the latent, the standardized $\tilde z = (z-\mu)/\sigma$ and the unit-normalized $\bar{z} = z/\|z\|$, and two health indicators. The latent compass is the pair of polar coordinates of $z_t$ in the anchor plane,
\begin{equation}
\theta_t = \arctan2\!\bigl(z_t\cdot \mathbf{e}_2,\ z_t\cdot \mathbf{e}_1\bigr),\qquad r_t = \bigl\|(z_t\cdot\mathbf{e}_1,\ z_t\cdot\mathbf{e}_2)\bigr\|,
 \label{eq:latent_his}
\end{equation}
which uses only the two anchor dimensions It uses only the two anchor dimensions, so it is a visual summary. For $\dim(z)>2$ the anchor plane holds only 30--54\% of the latent norm. The full-dimensional readout is the Mahalanobis distance from the healthy state, $h_t = \sqrt{(z_t-\mu_0)^\top\Sigma_0^{-1}(z_t-\mu_0)}$, with $(\mu_0,\Sigma_0)$ estimated on the healthiest training windows (Appendix~\ref{appendix_geometric_interpretation}). Both are computed from $z_t$ alone at inference time.

\section{Experimental Framework}\label{sec:experiments}

\paragraph{Datasets.} \textbf{C-MAPSS}~\citep{Saxena_Goebel_Simon_Eklund_2008} has four turbofan run-to-failure subsets (FD001--004) with one or six operating conditions (OCs) and one or two failure modes (FMs) (Appendix Table~\ref{tab:dataset_details}); training units are complete, test units are right-censored with known RUL. \textbf{N-CMAPSS}~\citep{arias2021ncmapss} simulates turbofans under real flight profiles; we follow the protocol-matched CruiseBench pipeline~\citep{cheng2026cruisebench} (DS02-006, 9 units, cruise mask, window 256), with the RUL cap of 34 cycles that CruiseBench prescribes for DS02-006; results at caps 80 and 125 are in Appendix Table~\ref{tab:ncmapss-caps}. \textbf{NASA Li-ion battery aging}~\citep{saha2007battery}: the four classic cells B0005/6/7/18, evaluated leave-one-cell-out (Appendix C). For qualitative analysis we use NVIDIA GR00T's Can2Drawer~\citep{gr00tn1_2025} and LeRobot~\citep{cadene2024lerobot} manipulation episodes (joint positions), and a mouse neural-activity recording~\citep{Vinograd_Nair_Kim_Linderman_Anderson_2024} (Appendix~\ref{appx:mouse}). 
Sensor channels are min-max normalized per feature (after a per-regime z-score on FD002/FD004), with the scaler fit on training windows only, and the setting columns are not given to the encoder.

\paragraph{Protocol.} The backbone model is trained without RUL targets and frozen, then a downstream head is fit on its latent features $z$. On C-MAPSS the head is a transformer (Tr.) or a linear regressor (LR.) on $\tilde z$, $\theta$, $r$ and the progression score (Appendix~\ref{appendix_geometric_interpretation}). We report mean $\pm$ std over 5 seeds, RUL RMSE and the asymmetric PHM08 score~\citep{Saxena_Goebel_Simon_Eklund_2008}, and unit-level bootstrap 95\% confidence intervals in Appendix~\ref{appx:review_results}. HI quality is measured by \emph{trendability}, the mean over units of the absolute Spearman correlation $|\rho(\text{HI},\text{RUL})|$ (higher is better), with monotonicity and prognosability in the appendix~\citep{coble2010merging}.

\paragraph{Baselines.} We compare STEP with two variants of the same backbone that differ only in their training objective, BaseAE (reconstruction only) and SoftCLT~\citep{Lee_Park_Lee_2024}, using identical windows, features and RUL heads. For health indicators, we also build three classical HIs directly on the raw sensor features, without a trained encoder. These are the projection on the first principal component, the best-matching-unit index on a one-dimensional SOM, and a similarity indicator given by the Mahalanobis distance from the healthy state. On C-MAPSS, we further report published RUL models. On N-CMAPSS, we reproduce a supervised GRU and TSMixer~\citep{chen2023tsmixer} in the same pipeline, and add a randomly initialized encoder as a null baseline.


\section{Results and Discussion}\label{sec:discussion}

\subsection{Latent Geometry}\label{sec:geometry}

\begin{figure}[t]
    \centering
    \includegraphics[width=\textwidth]{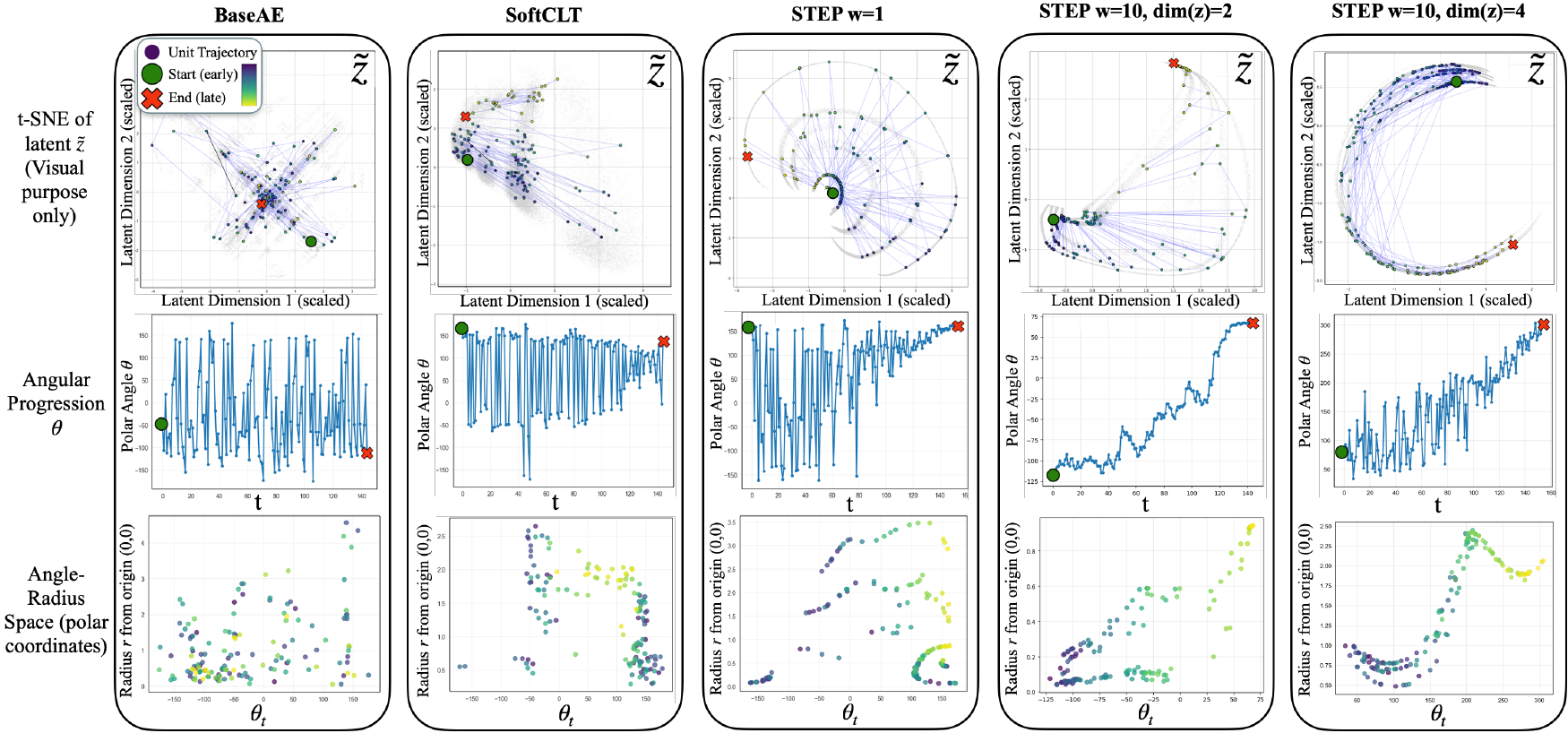}
    \caption{\textbf{Latent compass on C-MAPSS FD002} (6 operating conditions). Gray: t-SNE of the whole dataset; colored: engine 20 over time. AE shows no coherent structure and SoftCLT a limited one. STEP with window $w{=}1$ separates six concentric rings, one per operating condition, with $\theta$ advancing along the degradation. With $w{=}10$ the rings merge while the angular progression sharpens. With $\dim(z){=}4$ the embedding is richer and the rings less pronounced. The rings are a dataset- and setting-dependent visual effect; quantitatively, the radius is not a reliable mode label (Sec.~\ref{sec:discussion_theta_vs_t}).}
    \label{fig:latent_his_results}
\end{figure}

Fig.~\ref{fig:latent_his_results} compares the latent geometry of the three objectives on the same backbone. AE, trained for reconstruction only, leaves FD002 diffuse and temporally incoherent; SoftCLT forms soft clusters loosely aligned with OCs but no global ordering. STEP produces an ordered manifold between the anchors: at $w{=}1$ six rings appear, one per OC, and $\theta$ advances with degradation; at $w{=}10$ the rings merge and the progression sharpens. The 2D picture thus depends on window and dimension (FD004 merges at higher $\dim(z)$, FD003 keeps its separation), whereas RUL accuracy does not (Sec.~\ref{sec:ablations}), and the full-dimensional readout stays trendable at every dimension. Using $\theta$ and $r$ as diagnostics for AE and SoftCLT is motivated in Appendix~\ref{app:polar_diagnostics}. On robotics (Fig.~\ref{fig:robot_geometry}), $\bar{z}$ groups episodes into task phases (locating, grabbing, moving, placing) without annotation; this evidence is qualitative.

\begin{figure}[t]
    \centering
    \begin{subfigure}[b]{0.49\textwidth}
        \centering
        \includegraphics[width=0.72\linewidth]{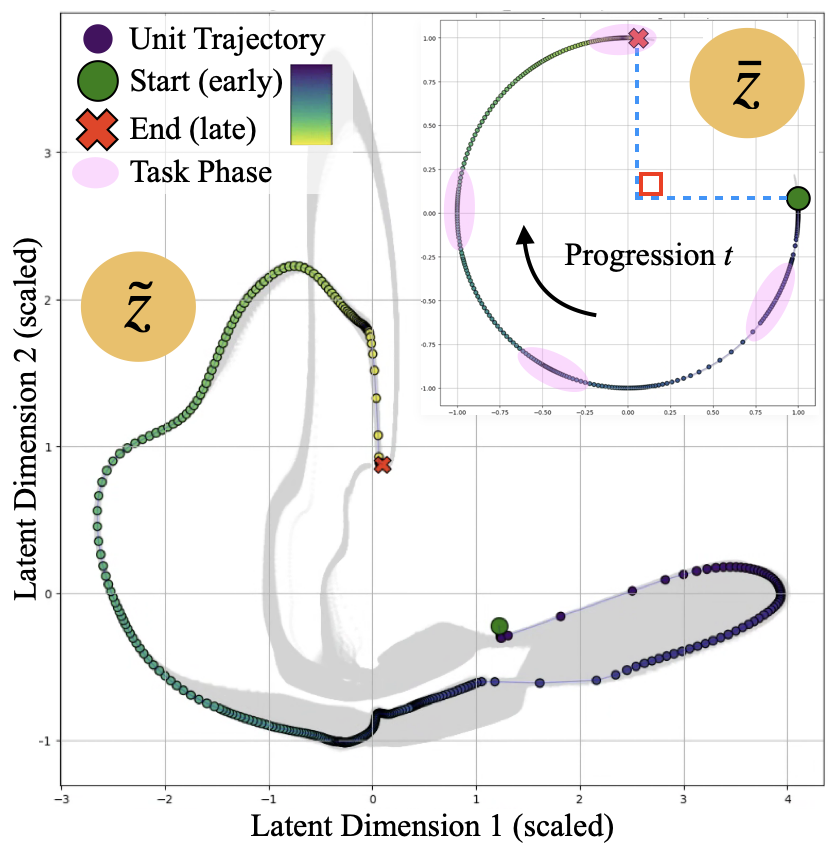}
        \caption{GR00T~\citep{gr00tn1_2025} (pick-and-place).}
        \label{fig:robot_groot}
    \end{subfigure}
    \hfill
    \begin{subfigure}[b]{0.49\textwidth}
        \centering
        \includegraphics[width=0.72\linewidth]{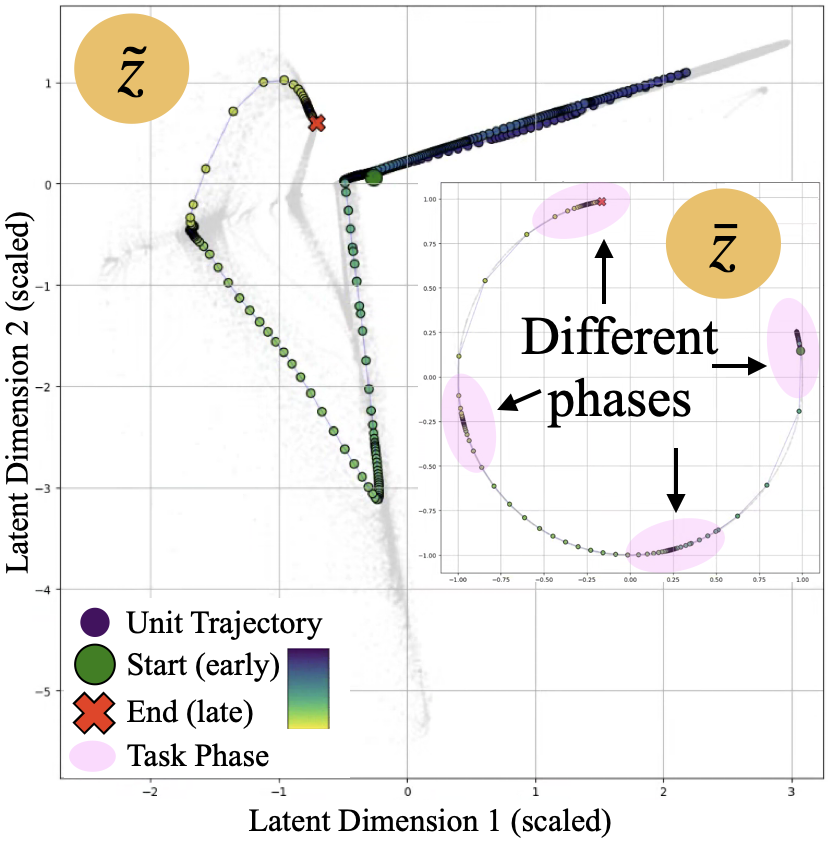}
        \caption{LeRobot~\citep{cadene2024lerobot} (shelf grabbing).}
        \label{fig:robot_lerobot}
    \end{subfigure}
    \caption{\textbf{STEP latents on two robotics datasets} (qualitative). $\tilde{z}$ follows task progression; $\bar{z}$ groups into task phases (locating, grabbing, moving, placing) across two embodiments and tasks.}
    \label{fig:robot_geometry}
\end{figure}

\subsection{Does $\theta$ Track State Beyond Elapsed Time?}
\label{sec:discussion_theta_vs_t}
A natural concern is that $\theta$ merely encodes the clock, since triplets are built from temporal proximity. We test this on \emph{held-out} units of the multi-condition subsets FD002 and FD004 with controls on elapsed cycles, not on $t/T$, which needs the oracle $T$ and carries no information about absolute RUL.

\emph{Remaining life beyond elapsed time.} If $\theta$ only encoded elapsed cycles, controlling for them would remove its relation to RUL. The partial correlation $|\rho(\theta,\mathrm{RUL}\mid t)|$ is 0.30 ($p<0.001$; mean over FD002 and FD004) for STEP, against 0.03 for the AE angle: the AE angle is close to a clock, STEP's is not.

\emph{Invariance to operating condition.} Within matched-$\theta$ bins, a 6-way OC classifier recovers the OC from the AE angle at balanced accuracy 0.52, but from STEP's angle only at chance (0.17; chance $=1/6$). STEP's progression axis is thus not confounded by the operating regime.

\emph{What the radius does and does not carry.} 
The operating condition is not lost, though. It stays partly readable from the radius (balanced accuracy 0.37 on FD002 and 0.39 on FD004) and almost fully from the full latent (0.97 and 0.98). STEP thus moves operating-condition information off the angle and into the other dimensions, which is what the rings of Fig.~\ref{fig:latent_his_results} show, even if the radius alone is not a reliable mode label. The full latent also separates failure modes. Clustering the terminal embeddings with K-means splits the two failure modes of FD003 (silhouette 0.75--0.97) and FD004 (0.60--0.66) without supervision, despite a single $z_\text{end}$ (Appendix Table~\ref{tab:kmeans}).

\subsection{RUL Estimation on C-MAPSS}\label{sec:rul}

\begin{table}[t]
\centering
\small
\caption{\textbf{RUL RMSE on C-MAPSS} (lower is better). HI marks methods that also provide health indicators. I-GLIDE derives them from data deemed healthy, whereas STEP maps the full trajectory from start to end of life. Top: published results as reported by their authors, typically single or best runs. Bottom: the same backbone trained with different objectives, frozen, with a transformer (Tr.) or linear (LR.) head, mean $\pm$ std over 5 seeds. Bold marks the best mean per subset. SoftCLT rows use the same backbone, features and heads as STEP, scored through a latent-to-head bridge that reproduces the published Tr.STEP values to within 0.4 RMSE. PHM08 scores and all SoftCLT latent sizes are in Appendix Tables~\ref{tab:seeds_phm} and~\ref{tab:softclt_full}.}
\label{tab:comparison_cmaps}
\setlength{\tabcolsep}{5pt}
\begin{tabular}{lccccc}
\toprule
\textbf{Model} & \textbf{HI} & \textbf{FD001} & \textbf{FD002} & \textbf{FD003} & \textbf{FD004} \\
\midrule
MLP~\citep{Zheng_Ristovski_Farahat_Gupta_2017}   & & 37.56 & 80.03 & 37.39 & 77.37 \\
SVR~\citep{Zheng_Ristovski_Farahat_Gupta_2017}   & & 20.96 & 42.00 & 21.05 & 45.35 \\
RVR~\citep{Zheng_Ristovski_Farahat_Gupta_2017}   & & 23.80 & 31.30 & 22.37 & 34.34 \\
CNN~\citep{Zheng_Ristovski_Farahat_Gupta_2017}   & & 18.45 & 30.29 & 19.82 & 29.16 \\
SSL~\citep{Listou_Ellefsen_Bjorlykhaug_AEsoy_Ushakov_Zhang_2019} & & 12.56 & 22.73 & 12.10 & 22.66 \\
CNN-LSTM~\citep{Peng_Chen_Chen_Tang_Li_Gui_2021} & & 11.17 & -- & 9.99 & -- \\
MS-DCNN~\citep{Li_Zhao_Zhang_Zio_2020}           & & 11.44 & 19.35 & 11.67 & 22.22 \\
VAE + RNN~\citep{Costa_Sanchez_2022}             & & 11.44 & 24.12 & 14.88 & 26.54 \\
MLE4X+CCF~\citep{Pillai_Vadakkepat_2021}         & & 11.57 & 18.84 & 11.83 & 20.78 \\
RVE~\citep{Costa_Sanchez_2022}                   & & 13.42 & 14.92 & 12.51 & 16.37 \\
Prob. CNN~\citep{De_Pater_Mitici_2022}           & & 12.42 & 13.72 & 12.16 & 15.95 \\
Wang et al.~\citep{Wang_Wang_Wang_Wang_2025}     & \checkmark & -- & -- & -- & 15.42 \\
I-GLIDE~\citep{thil2025glide}                    & \checkmark & \textbf{9.47} & 16.18 & \textbf{8.29} & \textbf{12.32} \\
\midrule
Tr.AE $z{=}32$      & & 17.15{\scriptsize$\pm$0.18} & 18.13{\scriptsize$\pm$0.21} & 20.72{\scriptsize$\pm$1.41} & 20.76{\scriptsize$\pm$0.68} \\
Tr.SoftCLT $z{=}4$  & & 15.34{\scriptsize$\pm$1.17} & 21.23{\scriptsize$\pm$0.98} & 14.64{\scriptsize$\pm$1.07} & 22.77{\scriptsize$\pm$1.27} \\
Tr.SoftCLT $z{=}32$ & & 15.50{\scriptsize$\pm$0.61} & 19.73{\scriptsize$\pm$1.25} & 14.87{\scriptsize$\pm$1.06} & 23.09{\scriptsize$\pm$2.08} \\
LR.SoftCLT $z{=}32$ & & 14.84{\scriptsize$\pm$0.36} & 16.85{\scriptsize$\pm$0.53} & 15.97{\scriptsize$\pm$3.11} & 19.08{\scriptsize$\pm$0.95} \\
Tr.STEP $z{=}2$     & \checkmark & 13.18{\scriptsize$\pm$0.79} & 12.13{\scriptsize$\pm$0.10} & 13.30{\scriptsize$\pm$0.46} & 15.31{\scriptsize$\pm$0.84} \\
Tr.STEP $z{=}4$     & \checkmark & 13.12{\scriptsize$\pm$0.23} & 11.78{\scriptsize$\pm$0.40} & 12.89{\scriptsize$\pm$0.73} & 13.31{\scriptsize$\pm$0.32} \\
Tr.STEP $z{=}32$    & \checkmark & 13.27{\scriptsize$\pm$0.38} & 11.57{\scriptsize$\pm$0.34} & 12.66{\scriptsize$\pm$0.51} & 13.13{\scriptsize$\pm$0.69} \\
LR.STEP $z{=}32$    & \checkmark & 13.09{\scriptsize$\pm$0.42} & \textbf{11.44}{\scriptsize$\pm$0.04} & 12.48{\scriptsize$\pm$0.48} & 13.26{\scriptsize$\pm$0.39} \\
\bottomrule
\end{tabular}
\end{table}

Table~\ref{tab:comparison_cmaps} reports five-seed results. On the multi-condition FD002, STEP obtains the lowest mean RMSE among compared methods (11.57 $\pm$ 0.34 at $z{=}32$, against 13.72 for the best published baseline). On FD004 it is competitive (13.13 $\pm$ 0.69) but behind I-GLIDE (12.32). SoftCLT, trained with the same backbone and scored with the same heads and features, is 5-8 RMSE behind STEP on FD002 at matched latent dimension (best: $16.85\pm0.53$ with a linear head at $z{=}32$, against $11.48\pm0.31$ for STEP through the same evaluation path), and widening its latent from $z{=}4$ to $z{=}32$ does not improve results. On the single-condition subsets FD001 and FD003, STEP is behind the best supervised models, so its main value there lies in the health indicators rather than in RMSE. I-GLIDE also provides health indicators, but it learns only from data it first identifies as healthy, while STEP learns from the whole trajectory. 

Replacing the transformer head with a linear regressor on the same features changes RMSE by at most 0.2 on every subset. A linear head can only succeed if the RUL-relevant state is linearly accessible in the representation. This is evidence that STEP's coordinates are predictive and geometrically accessible. With the same backbone, the reconstruction-only AE is 4 to 8 RMSE worse, so the gain comes from the learned representation through STEP's training objective. 

\subsection{Health-Indicator Quality Across Benchmarks}\label{sec:hi_quality}

\begin{table}[t]
\centering
\small
\caption{\textbf{Health-indicator trendability} (mean per-unit $|\rho(\text{HI},\text{RUL})|$, higher is better). \textbf{(a)} C-MAPSS, $z{=}32$, 5 seeds: same backbone trained with reconstruction only (AE) or with STEP, read with the 2D compass angle or the full-dimensional Mahalanobis readout. \textbf{(b)} NASA battery, held-out cells (leave-one-cell-out). \textbf{(c)} 2D compass versus full-dimensional readout for STEP.}
\label{tab:hi_quality}
\begin{subtable}[t]{0.52\textwidth}
\centering
\caption{C-MAPSS, AE vs.\ STEP}
\setlength{\tabcolsep}{3pt}
\begin{tabular}{lcccc}
\toprule
 & \multicolumn{2}{c}{2D compass} & \multicolumn{2}{c}{Full (Mahalanobis)} \\
 \cmidrule(lr){2-3}\cmidrule(lr){4-5}
 & AE & STEP & AE & STEP \\
\midrule
FD001 & .25{\scriptsize$\pm$.02} & .68{\scriptsize$\pm$.16} & .35{\scriptsize$\pm$.01} & .70{\scriptsize$\pm$.01} \\
FD002 & .34{\scriptsize$\pm$.01} & .54{\scriptsize$\pm$.14} & .40{\scriptsize$\pm$.01} & .65{\scriptsize$\pm$.03} \\
FD003 & .27{\scriptsize$\pm$.01} & .72{\scriptsize$\pm$.08} & .46{\scriptsize$\pm$.02} & .69{\scriptsize$\pm$.02} \\
FD004 & .37{\scriptsize$\pm$.06} & .64{\scriptsize$\pm$.02} & .44{\scriptsize$\pm$.01} & .70{\scriptsize$\pm$.02} \\
\bottomrule
\end{tabular}
\end{subtable}\hfill

\begin{subtable}[t]{0.22\textwidth}
\centering
\caption{Battery}
\setlength{\tabcolsep}{3pt}
\begin{tabular}{lc}
\toprule
HI & Trend. \\
\midrule
PCA  & 0.93 \\
Similarity & 0.97 \\
SOM & 0.76 \\
STEP & \textbf{0.95} \\
\bottomrule
\end{tabular}
\end{subtable}\hfill

\begin{subtable}[t]{0.24\textwidth}
\centering
\caption{2D vs.\ full}
\setlength{\tabcolsep}{3pt}
\begin{tabular}{lcc}
\toprule
 & 2D & Full \\
\midrule
C-MAPSS & .64 & .68{\scriptsize$\pm$.02} \\
N-CMAPSS & .66 & .93{\scriptsize$\pm$.02} \\
\bottomrule
\end{tabular}
\end{subtable}
\end{table}

\begin{table}[t]
\centering
\small
\caption{\textbf{N-CMAPSS} (CruiseBench protocol~\citep{cheng2026cruisebench}, window 256, 5 seeds). STEP encoders are trained with no RUL label, then frozen; a RUL head is fit on their features. Sup.: RUL is the encoder's training target. Trend.: HI trendability. Absolute RMSE is dominated by the RUL cap and is comparable only within this protocol.}
\label{tab:ncmapss}
\setlength{\tabcolsep}{5pt}
\begin{tabular}{lccc}
\toprule
Model & Sup. & RMSE & Trend. \\
\midrule
GRU, supervised (repr.)      & \checkmark & \textbf{1.81}{\scriptsize$\pm$0.37} & -- \\
TSMixer~\citep{chen2023tsmixer}, supervised (repr.) & \checkmark & 2.41{\scriptsize$\pm$0.07} & -- \\
\midrule
STEP, CNN encoder            & & 2.19{\scriptsize$\pm$0.08} & 0.91{\scriptsize$\pm$0.01} \\
STEP, GRU encoder            & & 2.30{\scriptsize$\pm$0.06} & \textbf{0.92}{\scriptsize$\pm$0.01} \\
STEP, LSTM encoder           & & 2.38{\scriptsize$\pm$0.06} & 0.89{\scriptsize$\pm$0.02} \\
STEP, PatchTST~\citep{nie2023patchtst} encoder & & 2.87{\scriptsize$\pm$0.21} & 0.86{\scriptsize$\pm$0.03} \\
\midrule
Reconstruction only (null)   & & 9.39{\scriptsize$\pm$1.32} & 0.06 \\
Random-init encoder (null)   & & 3.44{\scriptsize$\pm$0.53} & 0.14 \\
\bottomrule
\end{tabular}
\end{table}


\paragraph{Health indicators on C-MAPSS.} With the same backbone, reconstruction alone gives a weak indicator on C-MAPSS (trendability 0.35--0.46). Adding the STEP terms roughly doubles it on every subset (0.65--0.70, Table~\ref{tab:hi_quality}a), while also improving RUL accuracy (Table~\ref{tab:comparison_cmaps}). STEP also beats the strongest classical indicator, the similarity HI computed directly on the sensors, by +0.16 trendability over all of C-MAPSS (95\% CI $[0.15, 0.17]$). The gain is largest on the multi-condition subsets (+0.19 on FD002, +0.24 on FD004), smaller on FD001 (+0.05), and the two are tied on FD003 (Appendix Table~\ref{tab:hi_vs_similarity}).

\paragraph{Beyond C-MAPSS.} On held-out NASA battery cells (leave-one-cell-out, fixed end of life, no capacity-derived inputs; Appendix C), STEP reaches a trendability of 0.95, competitive with the similarity indicator (0.97) and PCA (0.93) and above SOM (0.76), close to raw capacity (0.98), which defines end of life here. On N-CMAPSS (Table~\ref{tab:ncmapss}), reconstruction alone gives unusable features (RMSE 9.4, trendability 0.06), and the STEP terms turn them into strong ones on all four encoder families (RMSE 2.2--2.9, trendability 0.86--0.92): the objective is not tied to one architecture. 
STEP does not reach the supervised GRU (1.81) but improves on the reproduced supervised TSMixer (2.41) at the benchmark cap; at caps 80 and 125 both supervised models lead STEP by 5--12\% (Appendix Table~\ref{tab:ncmapss-caps}). Unlike the supervised regressors, STEP also provides a trendable health indicator.

\paragraph{2D compass versus full readout.} The full-dimensional readout recovers signal that the 2D compass discards, most where health is a low-variance direction (N-CMAPSS 0.66 to 0.93); on C-MAPSS the compass already captures most of it (0.64 vs.\ 0.68). The compass is the visual summary, the Mahalanobis readout the quantitative one.

\subsection{Ablations and Robustness}\label{sec:ablations}

\paragraph{Loss terms.} The triplet term is what shapes the geometry. Without it, reconstruction alone orders the latent only weakly on C-MAPSS, or sometimes is completely disorganized and can't recover the underlying structure (states are unordered) and fails on N-CMAPSS (Tables~\ref{tab:hi_quality}a and~\ref{tab:ncmapss}). The prototype term helps most when trajectories span several regimes, as on FD004, and slightly constrains the homogeneous FD003 (Sec.~\ref{sec:method_proto}). On N-CMAPSS, its benefit depends on the encoder (Appendix~\ref{appx:review_results}). We do not ablate reconstruction, since removing it risks collapsing the latent, as it serves as a regularization term in preserving the content. 

\paragraph{Triplet sampling.} Mixing negatives from the same unit and from other units works best, and drawing negatives uniformly at random works worse. The positive window matters little on homogeneous subsets but a lot on FD004, where widening it raises trendability from 0.47 to 0.63. In practice, the negative margin should be about three times the positive one, and no per-dataset tuning was needed.



\paragraph{Dimension and window.} RUL accuracy stays stable as the latent grows from $\dim(z){=}2$ to 32 (Table~\ref{tab:comparison_cmaps}) and across window sizes. The settings that blur the rings in the 2D picture therefore cost no accuracy, and the downstream regressor benefits from a longer context window. Larger latents are harder to train, as trajectories scatter in the sparser space. The decaying prototype weight keeps training stable from $\dim(z){=}4$ upward (Appendix~\ref{appx:training}).

\paragraph{Sensor noise.} To test robustness, we retrain at $\dim(z){=}16$ with the sensor noise halved or multiplied by 2.5. Going from the lowest to the highest noise level, RMSE rises only moderately, from 11.34 to 13.20 on FD001 and from 14.23 to 15.12 on FD002 (single runs, Appendix Table~\ref{tab:noise}). Noisy readings do not break the progressive assumption, which concerns the hidden state rather than the sensors themselves. STEP’s representation is thus robust to measurement noise, not only to its own training conditions.

\section{Conclusion}\label{sec:conclusion}

We have presented STEP, a method for learning structured representations of progressive time series by framing state evolution as geometric paths in a contrastively learned latent space, anchored between two fixed prototypes. Health indicators can be read directly from this geometry, as a two-dimensional latent compass for visualization and a full-dimensional readout for quantitative use. On held-out units, the angle tracks the hidden state rather than elapsed time, and it separates progression from operating conditions, which remain available in the rest of the latent. Across degradation benchmarks, STEP roughly doubles the trendability of a reconstruction-only backbone, outperforms classical indicators on C-MAPSS and matches them on battery data. These gains do not come at the cost of accuracy, since a simple linear model on STEP's coordinates is competitive with deep architectures for RUL prediction. The objective transfers across four encoder families, and on robotic tasks the same geometry recovers task phases without annotation.

\paragraph{Limitations.} STEP is designed for progressive sequences and assumes that the underlying state does not go back. It therefore does not model reversible transitions such as maintenance, other interventions, or regime switching with recovery. Training also relies on complete run-to-failure trajectories, since both the sampler and the prototype term use their start and end, so censored units are left out, whereas survival objectives handle them natively. Finally, STEP separates operating conditions from progression but does not model how control inputs shape the degradation rate, which control-aware dynamics address~\citep{Zhao_Fink_2027}.

\paragraph{Future work.} A natural extension is to systems that return to earlier states, for example after maintenance repairs a degraded component. More broadly, the geometry STEP learns could serve as a map for system discovery, where intermediate stages emerge as new anchors, and learning how the system moves along this map, including under control inputs, would reveal the mechanisms that drive its progression.

\begin{ack}
This work was carried out within the IRT SystemX of smart industrial digital twins (JNI3) project and was granted access to the HPC resources of IDRIS under the allocation 2025-AD011017210 made by GENCI. No competing interests to be declared.
\end{ack}


\bibliographystyle{plainnat}
\bibliography{references_camera_ready}

\newpage
\appendix
\addtocontents{toc}{\protect\setcounter{tocdepth}{2}}

{
\hypersetup{linkcolor=black}
\renewcommand{\contentsname}{Appendix Contents}
\tableofcontents
}
\vspace{1em}

\section{Notation and Problem Formulation}\label{appx:notation}
\begin{table}[h]
  \caption{Notation used throughout the paper.}
  \label{tab:naming_conventions}
  \centering
  \small
  \begin{tabular}{ll}
    \toprule
      \textbf{Symbol} & \textbf{Description} \\
    \midrule
$x_t$ & Observation (window) at time $t$ \\
$s_t$ & Unobserved latent (e.g.\ health) state at time $t$ \\
$z_t$ & Learned latent: $z_t = f_{\text{enc}}(x_t)$ \\
$\tilde{z}$ & Standardized latent, $(z-\mu)/\sigma$ \\
$\bar{z}$ & Unit-normalized latent, $z/\|z\|$ \\
$\theta_t, r_t$ & Latent compass: angle and radius in the anchor plane $(\mathbf{e}_1,\mathbf{e}_2)$ (Eq.~\ref{eq:latent_his}) \\
$h_t$ & Full-dimensional readout: Mahalanobis distance of $z_t$ from the healthy state \\
$\hat{x}_t$ & Reconstruction: $\hat{x}_t = f_{\text{dec}}(z_t)$ \\
$(x_a, x_p, x_n)$, $(z_a, z_p, z_n)$, $(t_a, t_p, t_n)$ & Anchor, positive, negative samples, codes and time indices \\
$\delta_t$ & Temporal sampling margin \\
$z_{\text{init}}=\mathbf{e}_1, z_{\text{end}}=\mathbf{e}_2$ & Fixed prototype vectors \\
$X_{\text{init}}, X_{\text{end}}$ & First / last $k$ steps of each training unit \\
$\mathcal{L}_{\text{recon}}, \mathcal{L}_{\text{trip}}, \mathcal{L}_{\text{proto}}$ & Reconstruction, cosine triplet and prototype losses \\
    \bottomrule
  \end{tabular}
\end{table}

\paragraph{RUL formulation.} Each unit gives a trajectory $(x_1,\ldots,x_T)$, sampled independently and of varying length. The degradation dataset is $\mathcal{D} = \{(x^{(u)}_t, y^{(u)}_t = T_u-t)\}$, where $y$ is the RUL. We seek indicators computed from $x_t$ that are predictive of $y_t$.

\paragraph{Progressive latent Markov process.} The mechanism behind a progression can be written as a non-regressive Markov chain $\{s_t\}^{T_u}_{t=1}$. Without assuming a discrete state space, if discretized into $K\geq 2$ states $s_t\in\{1,\ldots,K\}$, we have $s_1\neq s_T$, $s_T$ is absorbing, and indices can be ordered so that $s_t\geq s_{t-1}$. STEP does not fit an explicit Markov model; this formulation motivates learning an ordered latent space.

\section{Model Architecture and Training}\label{appx:training}
The backbone is a transformer encoder with a small MLP decoder. The low-capacity decoder prevents the decoder from bypassing the latent structure (akin to posterior collapse in VAEs). The N-CMAPSS experiments replace the encoder by a CNN, GRU, LSTM or PatchTST~\citep{nie2023patchtst} encoder with the same objective. The C-MAPSS STEP autoencoder has 240{,}562 parameters; the N-CMAPSS encoder bodies have approximately 28k (CNN), 11k (GRU), 15k (LSTM) and 169k (PatchTST) parameters (counted from the architecture definitions, not logged).

\begin{table}[h]
\centering
\small
\caption{Backbone components.}
\label{tab:model_components}
\begin{tabular}{@{}p{0.20\textwidth} p{0.32\textwidth} p{0.40\textwidth}@{}}
\toprule
\textbf{Component} & \textbf{Description} & \textbf{Specification} \\
\midrule
Input projection & Linear projection & $\mathbb{R}^{\text{input\_dim}} \rightarrow \mathbb{R}^{d_\text{model}}$ \\
Transformer encoder & Stack of transformer layers & 4 layers, 8 heads, FF$=2{\times}d_\text{model}$, GELU, dropout 0.1 \\
Global pooling & Sequence aggregation & Adaptive average pooling \\
Latent projection & MLP to latent space & $\mathbb{R}^{d_\text{model}}\!\rightarrow\!\mathbb{R}^{32}\!\rightarrow\!\mathbb{R}^{\text{latent\_dim}}$ \\
Decoder & MLP reconstruction & $\mathbb{R}^{\text{latent\_dim}}\!\rightarrow\!\mathbb{R}^{64}\!\rightarrow\!\mathbb{R}^{128}\!\rightarrow\!\mathbb{R}^{\text{input\_dim}}$ \\
\bottomrule
\end{tabular}
\end{table}

\begin{table}[h!]
\centering
\small
\caption{Default training hyperparameters ($d_\text{model}=32$).}
\begin{tabular}{llll}
\toprule
\textbf{Parameter} & \textbf{Value} & \textbf{Parameter} & \textbf{Value} \\
\midrule
Window size $w$ & 10 & Triplet margin $m$ & 1.0 \\
Latent dimension & 2 (also 4, 8, 16, 32) & $\lambda_1$ (recon) & 1.0 \\
Batch size & 256 & $\lambda_2$ (triplet) & 1.0 \\
Learning rate & $1 \times 10^{-3}$ & $\lambda_3$ (prototype) & 0.1 $\rightarrow$ 0.05 \\
Seeds & 5 (probes: 3) & Prototype windows $k$ & 8 initial, 8 final \\
\bottomrule
\end{tabular}
\end{table}

\paragraph{Optimization and model selection.} The encoder is trained with AdamW (learning rate $10^{-3}$ with cosine decay, weight decay 0.01, batch 256) for 100 epochs, without early stopping. The transformer RUL head is trained on frozen features with Adam (learning rate $10^{-3}$, batch 32) for 20 epochs. The linear head is ordinary least squares.

\paragraph{Prototype-weight schedule.} The triplet weight is constant ($\lambda_2{=}1$). The prototype weight starts at $\lambda_3{=}0.1$ and decays geometrically by a factor $0.95$ per epoch to a floor of $0.05$, so anchoring mainly shapes the early geometry. (The submitted version called this ``$\beta$-scheduling''.)

\paragraph{Sampler.} Positives are drawn from the same unit as the anchor within $\delta_t{=}20$ cycles. Negatives are drawn at random from windows in a different coarse life band (remaining life ${>}125$, $75$--$125$, $30$--$75$, ${\le}30$ cycles), which lets valid pairs be sampled without scanning each trajectory. The prototype term uses the first and last $k{=}8$ windows of each training unit. SoftCLT is trained with its own augmentation-based objective (Appendix~\ref{appx:softclt}).

\subsection{Theoretical Basis for Polar-Coordinate Diagnostics}
\label{app:polar_diagnostics}
Using $\theta$ and $r$ as diagnostics follows the alignment/uniformity view of contrastive learning~\citep{wang2020understanding}: positive pairs should be close (alignment), and features should spread over the hypersphere (uniformity). For progressive time series, alignment appears as a smooth, monotone $\theta$, and uniformity as distinct states occupying distinct regions. The origin is invariant to the data distribution, and movement relative to it has a direction ($\theta$) and a magnitude ($r$). Methods not designed for circular geometry (AE, SoftCLT) can be read through the same lens: erratic $\theta/r$ indicates poor alignment or uniformity.

\subsection{Derived Indicators}\label{appendix_geometric_interpretation}
Besides $\theta$, $r$ (Eq.~\ref{eq:latent_his}) and $h_t$, the ablations use the angular monotonicity and a progression score:
\begin{equation}
\mathcal{M} = \frac{1}{T-1} \sum_{t=1}^{T-1} \mathbb{I}(\theta_{t+1} > \theta_t),\qquad
\hat{u} = \frac{z_T - z_1}{\|z_T-z_1\|},\quad s_i = z_i \cdot \hat{u},\quad s'_i = \frac{s_i - s_\text{min}}{s_\text{max}-s_\text{min}},
\end{equation}
and the line deviation $d_i = \|v_i - p_i\|$ with $u = z_T - z_1$, $v_i = z_i - z_1$, $p_i = \frac{v_i \cdot u}{\|u\|^2} u$.

\paragraph{HI metrics.} Trendability is the mean over units of $|\rho_\text{Spearman}(\text{HI}_t, \text{RUL}_t)|$. Monotonicity is $|\#\{\Delta\text{HI}>0\}-\#\{\Delta\text{HI}<0\}|/(T-1)$ per unit, and prognosability is $\exp\!\bigl(-\operatorname{std}(\text{HI}_\text{fail})/\operatorname{mean}|\text{HI}_\text{start}-\text{HI}_\text{fail}|\bigr)$, following~\citet{coble2010merging}.

\paragraph{Classical indicators.} All are computed on the raw sensor features of the last frame of each window, with no encoder. \emph{PCA-based}: projection on the first principal component. \emph{SOM-based}: index of the best-matching unit on a 1-D Kohonen chain of 24 units initialized along the first principal component. \emph{Similarity}: Mahalanobis distance from the healthy feature distribution. ``Healthy'' windows are the 15\% of training windows with the largest remaining life, the same reference as the STEP Mahalanobis readout.

\section{Datasets}
\begin{table}[h]
\caption{NASA C-MAPSS subsets.}
\centering
\small
\begin{tabular}{lcccc}
\toprule
 & FD001 & FD002 & FD003 & FD004 \\
\midrule
Train units & 100 & 260 & 100 & 248 \\
Test units & 100 & 259 & 100 & 249 \\
OCs & 1 & 6 & 1 & 6 \\
FMs & 1 & 1 & 2 & 2 \\
\bottomrule
\end{tabular}
\label{tab:dataset_details}
\end{table}

\paragraph{N-CMAPSS.} We use the CruiseBench~\citep{cheng2026cruisebench} protocol on DS02-006 (9 units): cruise mask, window 256 with stride 10, 18 channels, min-max scaling fit on the development units, RUL cap 34, the cap CruiseBench prescribes for DS02-006 (caps 80 and 125 in Table~\ref{tab:ncmapss-caps}). The supervised GRU and TSMixer are reproduced in the same pipeline. HI trendability in Table~\ref{tab:ncmapss} is the healthy-referenced Mahalanobis readout computed per test window; the 0.93 of Table~\ref{tab:hi_quality}c uses the same readout computed per unit with window 20.

\begin{table}[h]
\centering
\small
\caption{N-CMAPSS DS02-006 (CruiseBench protocol) RUL RMSE at caps 34 (benchmark cap), 80 and 125; mean $\pm$ sample std over 5 seeds. Each cap is a full rerun, since the cap also enters encoder training through the clipped remaining life. HI trendability at caps 34/80/125.}
\label{tab:ncmapss-caps}
\setlength{\tabcolsep}{4pt}
\begin{tabular}{lcccc}
\toprule
 & \multicolumn{3}{c}{RUL RMSE (cap)} & Trend. \\
\cmidrule(lr){2-4}
Model & 34 & 80 & 125 & 34/80/125 \\
\midrule
STEP, CNN         & 2.19{\scriptsize$\pm$0.08} & 6.62{\scriptsize$\pm$0.06} & 6.75{\scriptsize$\pm$0.10} & .91/.90/.88 \\
STEP, GRU         & 2.30{\scriptsize$\pm$0.06} & 6.84{\scriptsize$\pm$0.14} & 6.92{\scriptsize$\pm$0.10} & .92/.67/.68 \\
STEP, LSTM        & 2.38{\scriptsize$\pm$0.06} & 6.98{\scriptsize$\pm$0.20} & 6.90{\scriptsize$\pm$0.14} & .89/.85/.91 \\
STEP, PatchTST    & 2.87{\scriptsize$\pm$0.24} & 7.11{\scriptsize$\pm$0.35} & 7.20{\scriptsize$\pm$0.07} & .86/.82/.86 \\
\midrule
GRU, supervised     & 1.81{\scriptsize$\pm$0.37} & 6.33{\scriptsize$\pm$0.52} & 6.36{\scriptsize$\pm$0.30} & -- \\
TSMixer, supervised & 2.41{\scriptsize$\pm$0.08} & 6.18{\scriptsize$\pm$0.22} & 6.03{\scriptsize$\pm$0.10} & -- \\
\midrule
Recon-only (null)   & 9.39{\scriptsize$\pm$1.32} & 18.13{\scriptsize$\pm$2.44} & 16.87{\scriptsize$\pm$3.10} & .06/.09/.09 \\
Random-init (null)  & 3.44{\scriptsize$\pm$0.53} & 10.01{\scriptsize$\pm$2.71} & 9.66{\scriptsize$\pm$2.07} & .14/.15/.15 \\
\bottomrule
\end{tabular}
\end{table}

\paragraph{NASA battery.} We evaluate the four classic cells B0005, B0006, B0007 and B0018 with leave-one-cell-out cross-validation (train on three cells, test on the held-out one). End of life is fixed at 1.4 Ah, and each trajectory is clipped at the first cycle below this threshold. Each discharge cycle is represented by its voltage and temperature curves, resampled on a fixed 64-point time grid and truncated at the shortest training-cell discharge, so that neither capacity nor discharge duration is given to the model. The classical indicators receive the same input, and raw capacity serves as a reference. The encoder is a GRU (dim(z)=8). Its hidden size and number of epochs are selected by an inner leave-one-cell-out loop on the training cells only, and the held-out cell is scored once. The sampler bands and the healthy reference use the state of health (capacity over its initial maximum) of the training cells only. We report trendability on the held-out cell, mean ± sample std over 4 folds × 3 seeds. In an earlier version that also gave the encoder discharge duration, trendability was inflated, because at constant 2 A current duration is proportional to capacity.

\section{Additional Results}\label{appx:review_results}

\subsection{Five-Seed C-MAPSS Results, with PHM08}
\begin{table}[h]
\centering
\small
\caption{Tr.STEP on C-MAPSS over 5 seeds: mean $\pm$ sample std of RUL RMSE and PHM08 score (lower is better).}
\label{tab:seeds_phm}
\setlength{\tabcolsep}{5pt}
\begin{tabular}{lccc}
\toprule
Subset & $\dim(z)$ & RMSE (mean $\pm$ std) & PHM08 \\
\midrule
FD001 & 2  & 13.18 $\pm$ 0.79 & 193 $\pm$ 30 \\
FD001 & 4  & 13.12 $\pm$ 0.23 & 181 $\pm$ 13 \\
FD001 & 32 & 13.27 $\pm$ 0.38  & 179 $\pm$ 11 \\
\midrule
FD002 & 2  & 12.13 $\pm$ 0.10 & 340 $\pm$ 13 \\
FD002 & 4  & 11.78 $\pm$ 0.40 & 305 $\pm$ 12 \\
FD002 & 32 & 11.57 $\pm$ 0.34 & 288 $\pm$ 12 \\
\midrule
FD003 & 2  & 13.30 $\pm$ 0.46 & 258 $\pm$ 25 \\
FD003 & 4  & 13.66 $\pm$ 0.89 & 265 $\pm$ 45 \\
FD003 & 32 & 12.66 $\pm$ 0.51 & 263 $\pm$ 39 \\
\midrule
FD004 & 2  & 15.31 $\pm$ 0.84 & 688 $\pm$ 105 \\
FD004 & 4  & 13.31 $\pm$ 0.32 & 585 $\pm$ 20 \\
FD004 & 32 & 13.13 $\pm$ 0.69 & 563 $\pm$ 103 \\
\bottomrule
\end{tabular}
\end{table}
\subsection{STEP versus the Classical Similarity Indicator on C-MAPSS}
\begin{table}[h]
\centering
\small
\caption{Trendability of the STEP Mahalanobis readout versus the input-space similarity HI, 3 seeds, paired trajectory bootstrap ($B{=}4000$). This harness scores trajectories differently from Table~\ref{tab:hi_quality}a, so absolute values are not comparable across the two tables.}
\label{tab:hi_vs_similarity}
\begin{tabular}{lrccc}
\toprule
Subset & $n_{\text{traj}}$ & STEP [95\% CI] & Similarity HI [95\% CI] & $\Delta$ [95\% CI] \\
\midrule
Pooled & 709 & 0.809\,[0.796,\,0.821] & 0.647\,[0.635,\,0.659] & $+0.162$\,[0.151,\,0.173] \\
FD001 & 100 & 0.845\,[0.810,\,0.878] & 0.794\,[0.775,\,0.812] & $+0.051$\,[0.026,\,0.075] \\
FD002 & 260 & 0.861\,[0.854,\,0.868] & 0.671\,[0.658,\,0.684] & $+0.190$\,[0.179,\,0.200] \\
FD003 & 100 & 0.664\,[0.604,\,0.721] & 0.665\,[0.627,\,0.700] & $-0.001$\,[$-0.040$,\,0.036] \\
FD004 & 249 & 0.798\,[0.781,\,0.815] & 0.556\,[0.532,\,0.578] & $+0.243$\,[0.230,\,0.256] \\
\bottomrule
\end{tabular}
\end{table}

\subsection{Held-Out Controls for $\theta$ and the Radius}
\begin{table}[h]
\centering
\small
\caption{Held-out C-MAPSS units. Top: partial correlation of the angle with RUL controlling for elapsed cycles. Bottom: 6-way OC classification, balanced accuracy (chance $0.17$); the angle probe controls for progression level.}
\label{tab:heldout_probes}
\begin{tabular}{lcc}
\toprule
 & AE & STEP \\
\midrule
$|\rho(\theta,\mathrm{RUL}\mid t)|$ & 0.03 & 0.30 ($p<0.001$) \\
OC from angle $\theta$ & 0.52 & 0.17 \\
\midrule
 & FD002 & FD004 \\
\midrule
OC from full STEP latent (linear probe) & 0.97 & 0.98 \\
OC from STEP radius $r$ & 0.37 & 0.39 \\
\bottomrule
\end{tabular}
\end{table}
Held-out units of FD002 and FD004, the partial correlation is the mean over the two subsets (STEP $-0.32$ and $-0.29$; AE $0.02$ and $0.03$). The angle probe measures balanced accuracy within matched-$\theta$ bins.

\subsection{Loss-Term and Sampling Ablations}
\paragraph{Loss terms.} C-MAPSS reconstruction-only versus full STEP is in Table~\ref{tab:hi_quality}a. On N-CMAPSS, reconstruction alone gives unusable features while the full objective gives strong ones on every encoder family (Table~\ref{tab:ncmapss}).

\paragraph{Triplet sampling.} Table~\ref{tab:triplet_sampling} varies the temporal-proximity threshold $\tau$ for positives and the negative mode. This harness uses a different RUL head from Table~\ref{tab:comparison_cmaps}, so its RMSEs are only comparable within the table.
\begin{table}[h]
\centering
\small
\caption{Triplet-sampling ablation, STEP, 3 seeds, FD001 and FD004. Trendability (anchor-plane angle) / RUL RMSE.}
\label{tab:triplet_sampling}
\begin{tabular}{llcc}
\toprule
$\tau$ & Negatives & FD001 & FD004 \\
\midrule
4  & mixed  & 0.634 / 17.03 & 0.471 / 23.74 \\
8  & mixed  & 0.638 / 17.19 & 0.511 / 23.43 \\
12 & mixed  & 0.639 / 17.19 & 0.572 / 21.70 \\
20 & mixed  & 0.659 / 17.15 & 0.628 / 20.44 \\
40 & mixed  & 0.695 / 17.14 & 0.579 / 21.05 \\
12 & random & 0.664 / 17.60 & 0.564 / 22.92 \\
12 & cross-unit & 0.667 / 17.61 & 0.507 / 23.29 \\
12 & far    & 0.616 / 17.44 & 0.586 / 22.34 \\
\bottomrule
\end{tabular}
\end{table}

\subsection{Prototype Ablation ($\lambda_3=0$)}\label{appx:proto_ablation}
Removing prototype anchoring degrades the heterogeneous FD004 (12.10 $\rightarrow$ 19.01 RMSE, $+57\%$) and slightly improves the homogeneous FD003 (13.35 $\rightarrow$ 11.59). These are single runs from the submitted version, at $\dim(z){=}4$. 

\subsection{SoftCLT Baseline}\label{appx:softclt}
\begin{table}[h]
\centering
\small
\caption{SoftCLT on C-MAPSS, same backbone, features and heads as STEP, 5 seeds (mean $\pm$ sample std RMSE). $^\dagger$One of five seeds diverged. PHM08 is 220--4400 for SoftCLT against 174--536 for STEP, since late predictions are penalised exponentially.}
\label{tab:softclt_full}
\setlength{\tabcolsep}{4pt}
\begin{tabular}{llcccc}
\toprule
Head & $z$ & FD001 & FD002 & FD003 & FD004 \\
\midrule
Transformer & 2  & 14.57{\scriptsize$\pm$1.13} & 22.75{\scriptsize$\pm$7.14}$^\dagger$ & 14.18{\scriptsize$\pm$0.65} & 21.93{\scriptsize$\pm$0.94} \\
Transformer & 4  & 15.34{\scriptsize$\pm$1.17} & 21.23{\scriptsize$\pm$0.98} & 14.64{\scriptsize$\pm$1.07} & 22.77{\scriptsize$\pm$1.27} \\
Transformer & 32 & 15.50{\scriptsize$\pm$0.61} & 19.73{\scriptsize$\pm$1.25} & 14.87{\scriptsize$\pm$1.06} & 23.09{\scriptsize$\pm$2.08} \\
\midrule
Linear & 2  & 15.19{\scriptsize$\pm$1.73} & 22.78{\scriptsize$\pm$7.13}$^\dagger$ & 15.02{\scriptsize$\pm$1.22} & 21.04{\scriptsize$\pm$0.70} \\
Linear & 4  & 15.57{\scriptsize$\pm$0.73} & 19.30{\scriptsize$\pm$0.56} & 14.54{\scriptsize$\pm$1.72} & 21.79{\scriptsize$\pm$0.46} \\
Linear & 32 & 14.84{\scriptsize$\pm$0.36} & 16.85{\scriptsize$\pm$0.53} & 15.97{\scriptsize$\pm$3.11}$^\dagger$ & 19.08{\scriptsize$\pm$0.95} \\
\bottomrule
\end{tabular}
\end{table}

\subsection{Multi-Mode Failure Discovery}

A single $z_\text{end}$ prototype could in principle merge qualitatively different terminal states. Clustering the terminal embeddings with unsupervised K-means ($k{=}2$) instead recovers the two failure modes of FD003 and FD004 without supervision (Table~\ref{tab:kmeans}).



\begin{table}[h]
\centering
\small
\caption{Unsupervised K-means ($k{=}2$) on terminal STEP embeddings. Distinct failure modes are recovered without supervision despite a single $z_\text{end}$ prototype.}
\label{tab:kmeans}
\begin{tabular}{lcccc}
\toprule
Dataset & $\dim(z)$ & Silhouette & Separation ratio & Misplaced (\%) \\
\midrule
FD003 & 4  & 0.75 & 6.3$\times$  & 0\% \\
FD003 & 16 & 0.93 & 21.0$\times$ & 0\% \\
FD003 & 32 & 0.97 & 52.8$\times$& 0\% \\
FD004 & 8  & 0.60 & 3.6$\times$  & 0\% \\
FD004 & 16 & 0.62 & 1.9$\times$ & 0\% \\
FD004 & 32 & 0.66 & 3.5$\times$  & 0\% \\
\bottomrule
\end{tabular}
\end{table}

\begin{figure}[h]
    \centering
    \begin{subfigure}[b]{0.32\textwidth}
        \includegraphics[width=\linewidth]{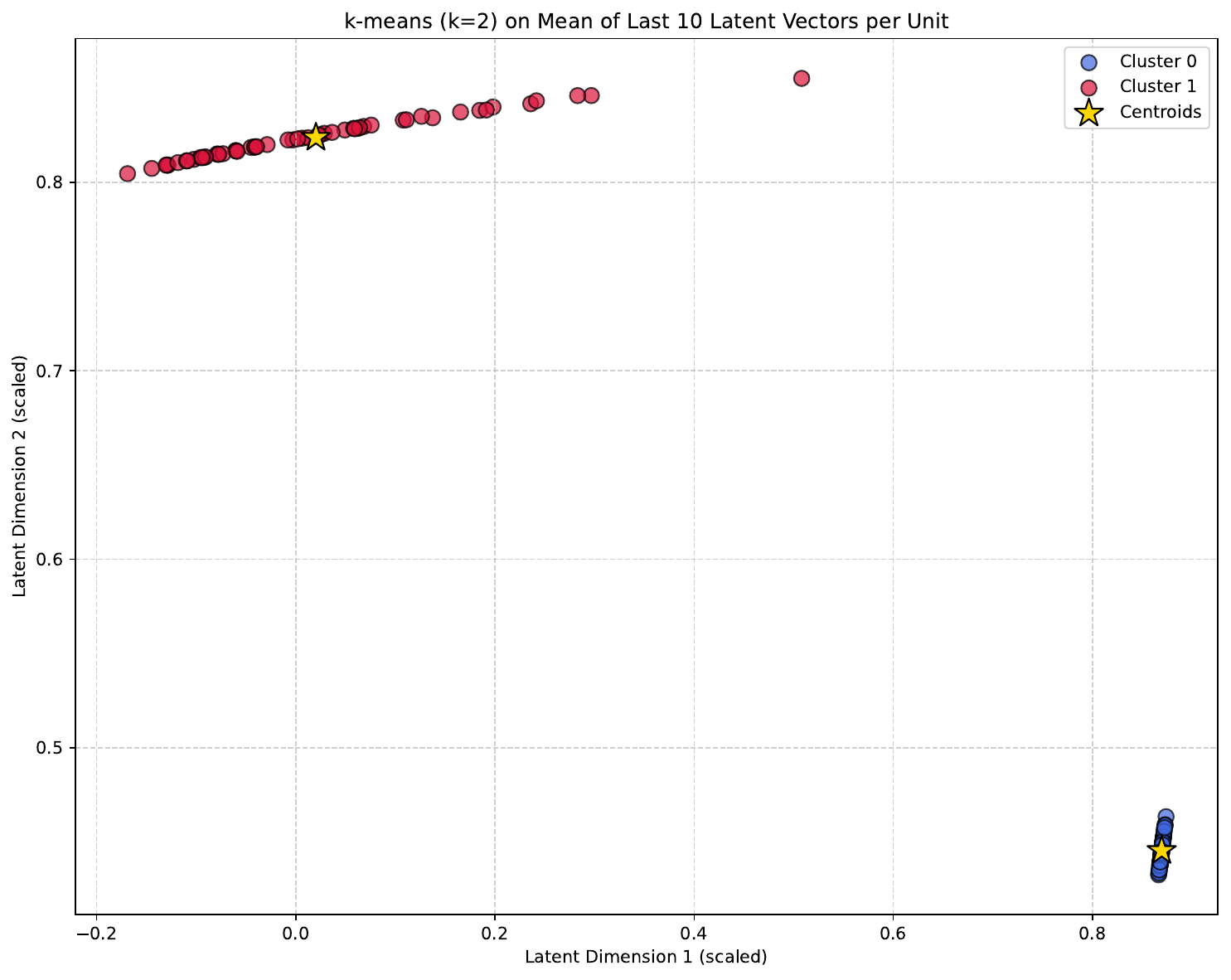}
        \caption{$\dim(z){=}4$.}
    \end{subfigure}
    \hfill
    \begin{subfigure}[b]{0.32\textwidth}
        \includegraphics[width=\linewidth]{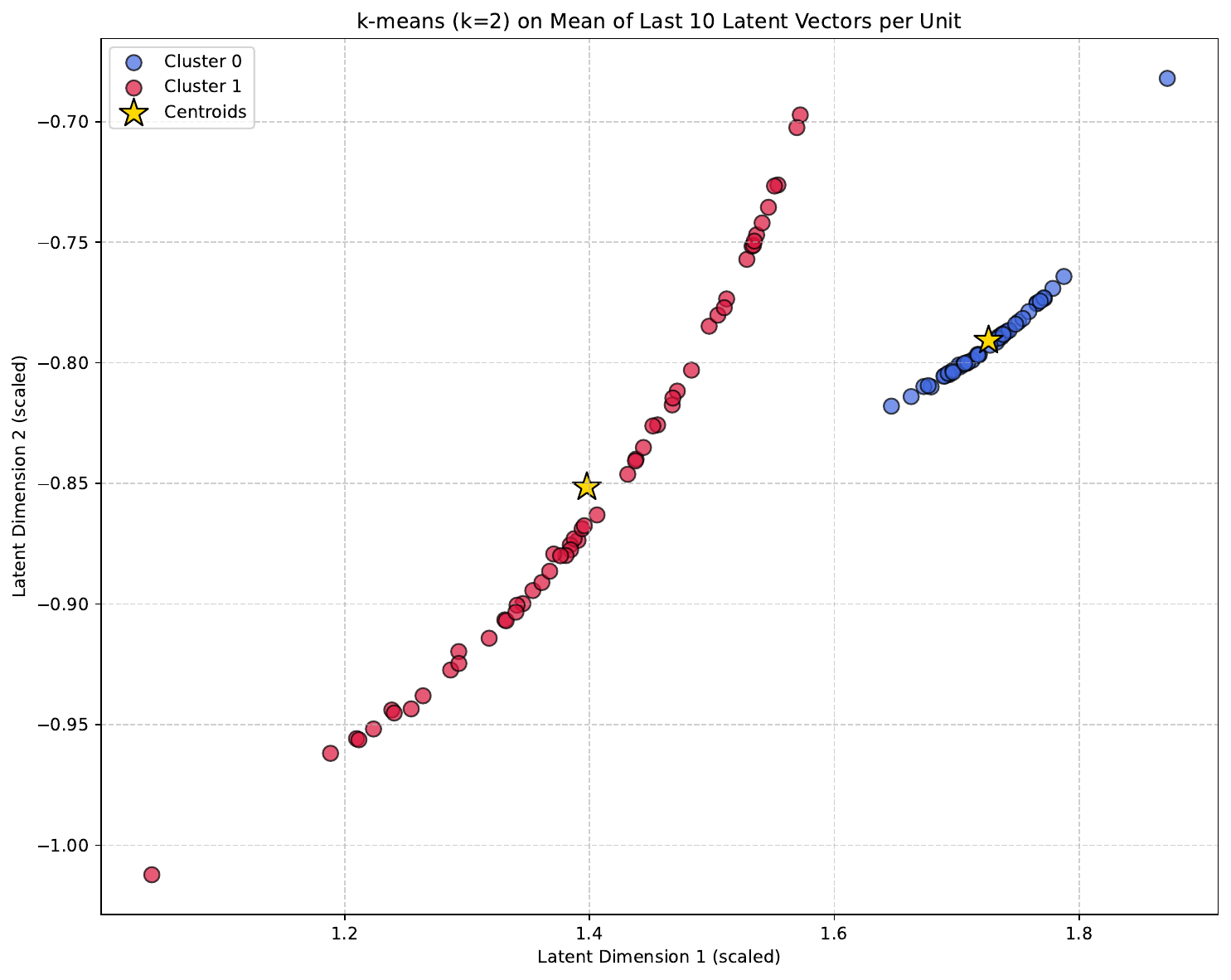}
        \caption{$\dim(z){=}16$.}
    \end{subfigure}
    \hfill
    \begin{subfigure}[b]{0.32\textwidth}
        \includegraphics[width=\linewidth]{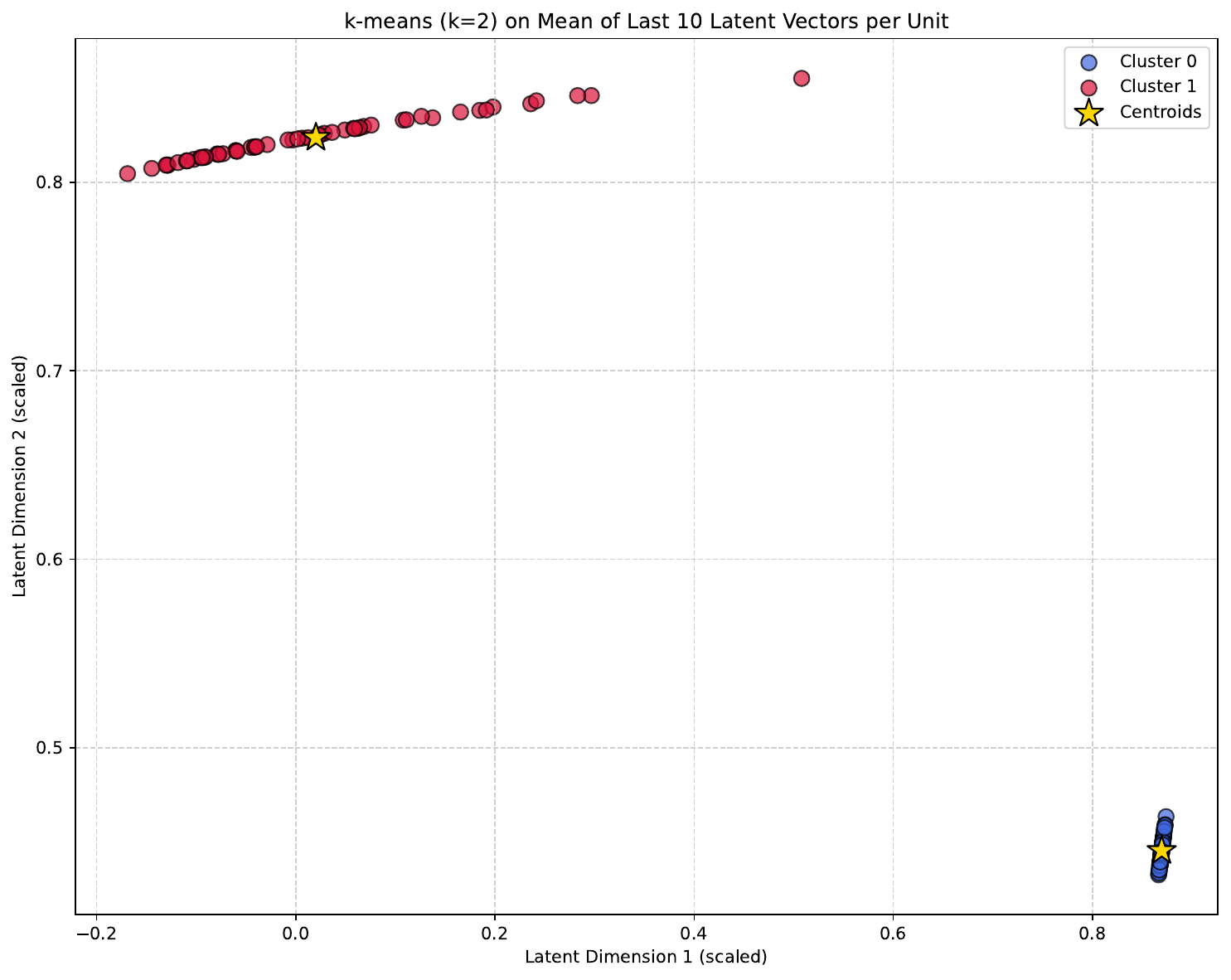}
        \caption{$\dim(z){=}32$.}
    \end{subfigure}
    \caption{\textbf{K-means ($k{=}2$) on terminal latent embeddings (mean of last 10 steps) recovers the two FD003 failure modes} at every latent dimension.}
\end{figure}

\subsection{Sensor-Noise Robustness}

We rescale the sensor noise by $\times0.5$ and $\times2.5$ and retrain. RMSE degrades gracefully (Table~\ref{tab:noise}), since non-regression is assumed of the latent state, not of the raw sensors.

\begin{table}[h]
\centering
\small
\caption{RUL RMSE at $\dim(z){=}16$ with sensor noise rescaled (single runs).}
\label{tab:noise}
\begin{tabular}{lcccc}
\toprule
Noise & FD001 & FD002 & FD003 & FD004 \\
\midrule
$\times 0.5$ & 11.34 & 14.23 & 13.23 & 18.36 \\
$\times 2.5$ & 13.20 & 15.12 & 14.52 & 24.70 \\
\bottomrule
\end{tabular}
\end{table}

\subsection{Feature Ablation for the RUL Head}

We vary the features given to the RUL head (Table~\ref{tab:ablation_study_z2}).

\begin{table}[htbp]
\centering
\small
\caption{Feature combinations for the RUL head, $\dim(z){=}2$, single runs from the submitted version, no feature smoothing, observation window 75. Bold: best per subset.}
\label{tab:ablation_study_z2}
\begin{tabular}{llcccc}
\toprule
\textbf{Model} & \textbf{Features} & \textbf{FD001} & \textbf{FD002} & \textbf{FD003} & \textbf{FD004} \\
\midrule
Random Forest & PS, $\theta$ & 14.70 & 12.90 & 13.43 & 18.02 \\
Random Forest & PS, $\theta$, $r$ & 13.04 & 13.04 & 12.98 & 18.13 \\
Random Forest & $\theta$, $r$ & 12.97 & 12.98 & 13.17 & 18.18 \\
Random Forest & $\theta$ & 14.41 & 12.86 & 13.37 & 18.67 \\
Random Forest & $z$ & 13.63 & 13.13 & 12.59 & 17.98 \\
Random Forest & $z$, PS, $\theta$, $r$ & 13.04 & 12.89 & 12.84 & 18.20 \\
Linear regression & $z$, PS, $\theta$, $r$ & 15.30 & 14.38 & 17.96 & 19.48 \\
Transformer & $z$ & 11.67 & 13.03 & \textbf{11.48} & 17.11 \\
Transformer & $\theta$, $z$ & \textbf{10.65} & 13.19 & 13.19 & 17.53 \\
Transformer & PS, $\theta$ & 14.12 & \textbf{12.57} & 13.93 & 17.44 \\
Transformer & PS, $\theta$, $r$ & 13.36 & 12.87 & 12.07 & \textbf{16.93} \\
Transformer & $\theta$ & 14.04 & 13.00 & 16.10 & 17.62 \\
Transformer & $\theta$, $r$ & 12.89 & 12.61 & 12.16 & 17.43 \\
Transformer & $z$, PS, $\theta$, $r$ & 12.73 & 12.66 & 12.02 & 17.08 \\
\bottomrule
\end{tabular}
\end{table}
Performance varies little across feature sets. I-GLIDE, by contrast, needed to combine several HIs and was unstable when they were used in isolation.

\section{Supplementary Diagnostics}
\label{appx:dim_sweep}

\subsection{$\theta$ versus Normalized Time on FD002}
The figures below are descriptive. Normalized time $t/T$ needs the oracle end of life and carries no information about absolute RUL, so comparisons against it are weaker than the held-out controls on elapsed cycles (Table~\ref{tab:heldout_probes}), which we consider the primary evidence.

\begin{figure}[h]
    \centering
    \includegraphics[width=0.95\textwidth]{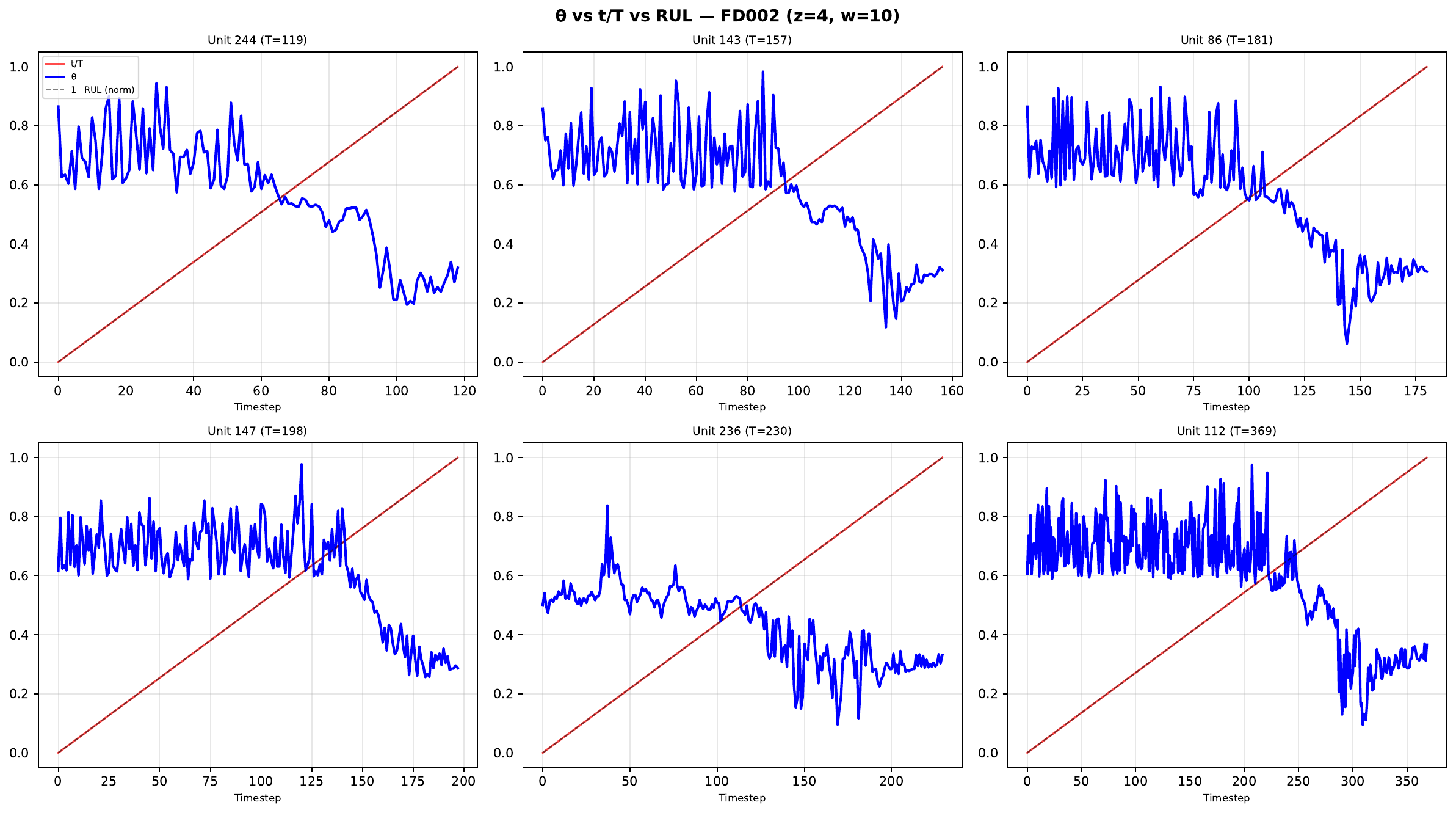}
    \caption{\textbf{Per-unit $\theta$ (blue) and $t/T$ (red) on six FD002 engines.} $t/T$ is a diagonal by construction; $\theta$ stays in a high, noisy healthy regime and then drops near failure, with transition timing that varies across engines.}
\end{figure}

\begin{figure}[h]
    \centering
    \includegraphics[width=0.6\textwidth]{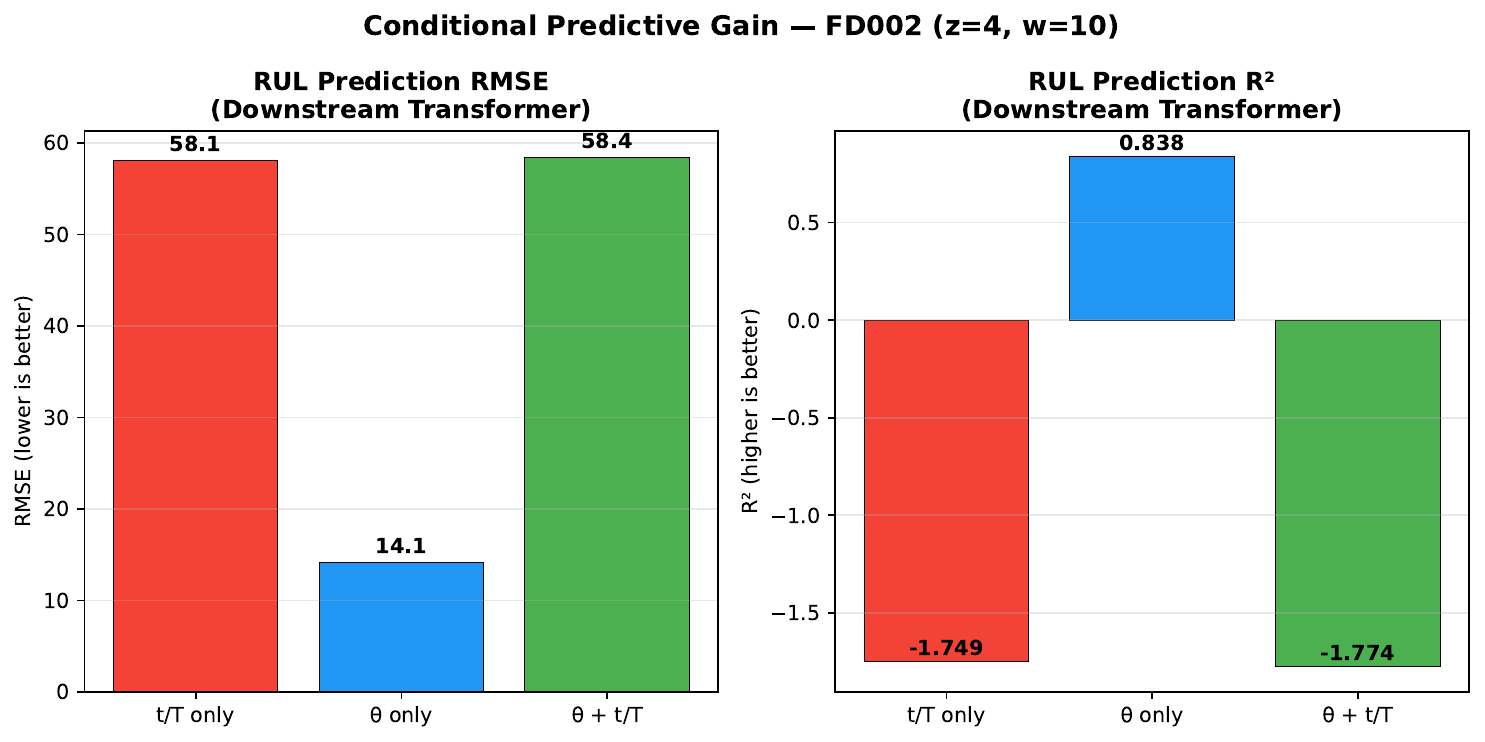}
    \caption{\textbf{Downstream RUL from $\theta$ versus $t/T$ (FD002).} A head on $\theta$ alone reaches RMSE 14.1 ($R^2{=}0.84$), one on $t/T$ alone 58.1; adding $t/T$ to $\theta$ does not help. Since $t/T$ carries no absolute-scale information, this shows that $\theta$ is predictive, not by itself that it is decoupled from elapsed time.}
    \label{fig:appx_predictive_gain}
\end{figure}

\begin{figure}[h]
    \centering
    \includegraphics[width=0.85\textwidth]{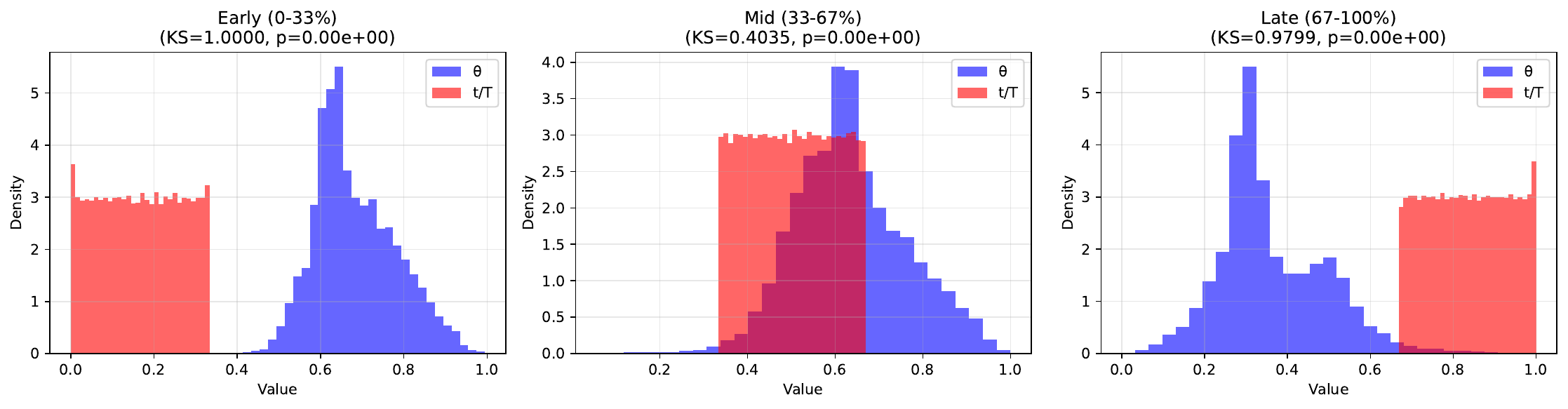}
    \caption{\textbf{Phase-wise marginal distributions of $\theta$ and $t/T$ on FD002} (KS$=$1.00 early, 0.40 mid, 0.98 late). Marginal differences do not by themselves test functional decoupling.}
    \label{fig:appx_phase_analysis}
\end{figure}

\begin{figure}[h]
    \centering
    \includegraphics[width=0.85\textwidth]{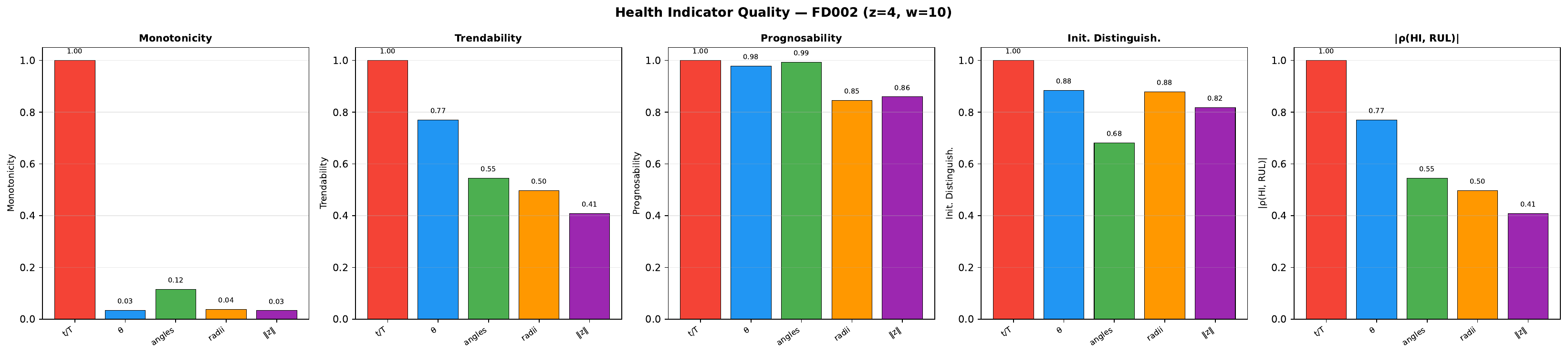}
    \caption{Standard PHM HI metrics on FD002. $t/T$ scores 1.0 on monotonicity and trendability by construction; $\theta$ reaches prognosability 0.98 while being computable at test time.}
\end{figure}

\subsection{Latent Dimension}
Note that the 2D compass uses only the anchor plane at every dimension; the per-dimension evidence below concerns the stability of that plane's geometry, and Table~\ref{tab:hi_quality}c the full-dimensional readout.

\begin{figure}[h]
    \centering
    \begin{subfigure}[b]{0.32\textwidth}
        \includegraphics[width=\linewidth]{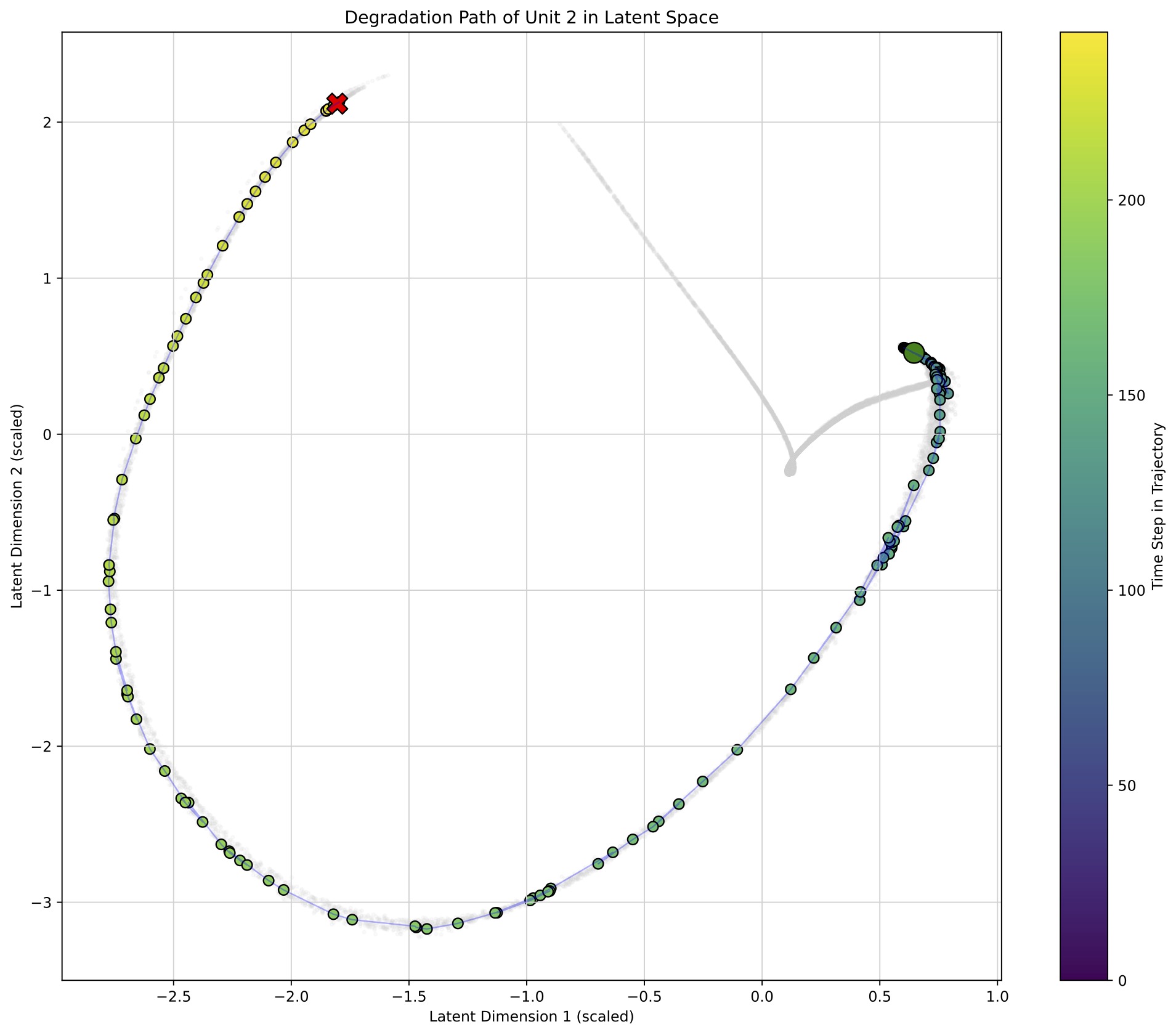}
        \caption{$\dim(z){=}4$.}
    \end{subfigure}
    \hfill
    \begin{subfigure}[b]{0.32\textwidth}
        \includegraphics[width=\linewidth]{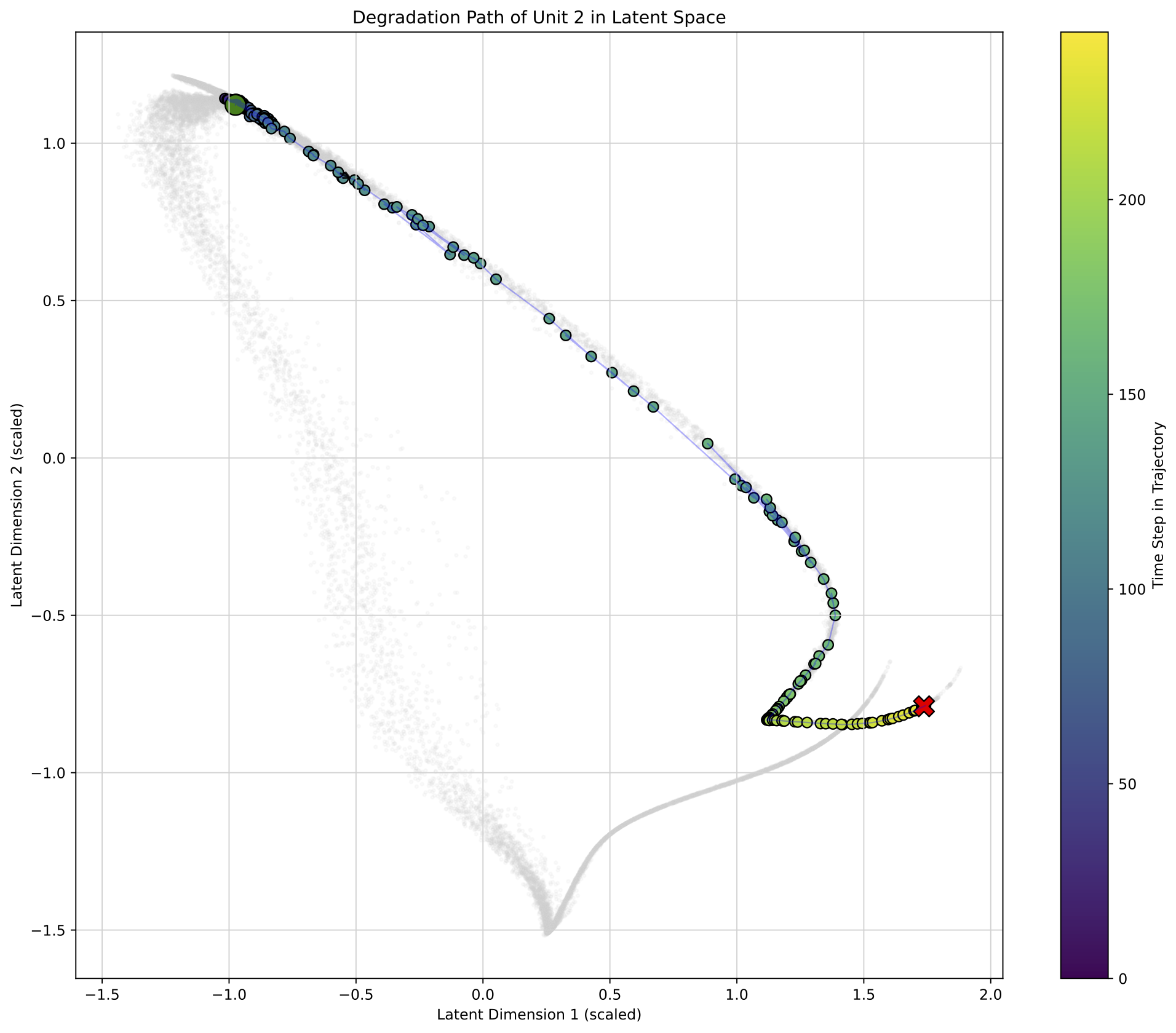}
        \caption{$\dim(z){=}16$.}
    \end{subfigure}
    \hfill
    \begin{subfigure}[b]{0.32\textwidth}
        \includegraphics[width=\linewidth]{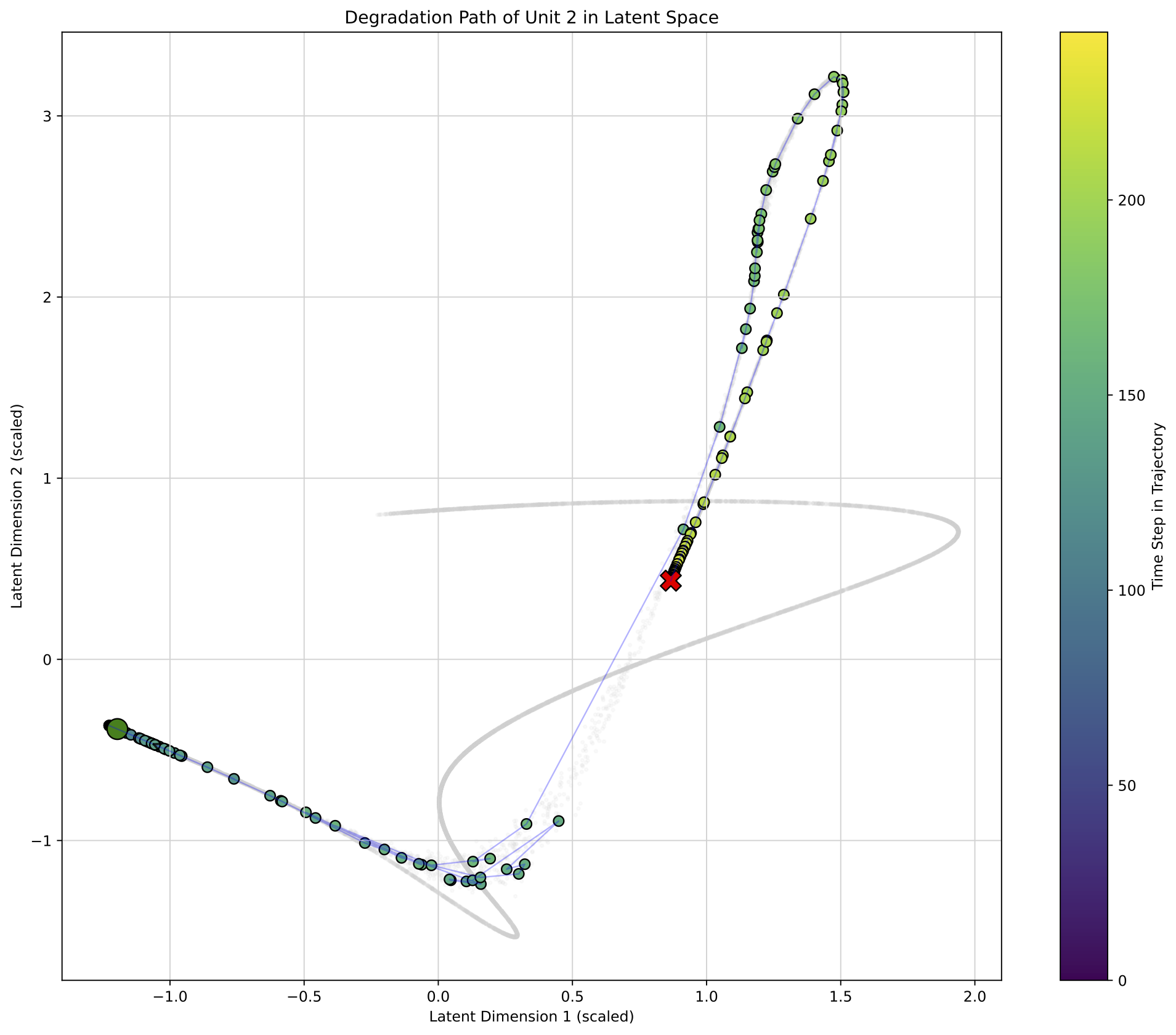}
        \caption{$\dim(z){=}32$.}
    \end{subfigure}
    \caption{\textbf{Anchor-plane geometry on FD003 across latent dimensions.} The direction from the healthy (green) to the end-of-life (red) anchor is preserved; per-unit $|\rho(\theta,\mathrm{RUL})|$ is 0.98 / 0.93 / 0.96 .}
    \label{fig:appx_dim_sweep_full}
\end{figure}

\begin{figure}[h]
    \centering
    \begin{subfigure}[b]{0.49\textwidth}
        \includegraphics[width=\linewidth]{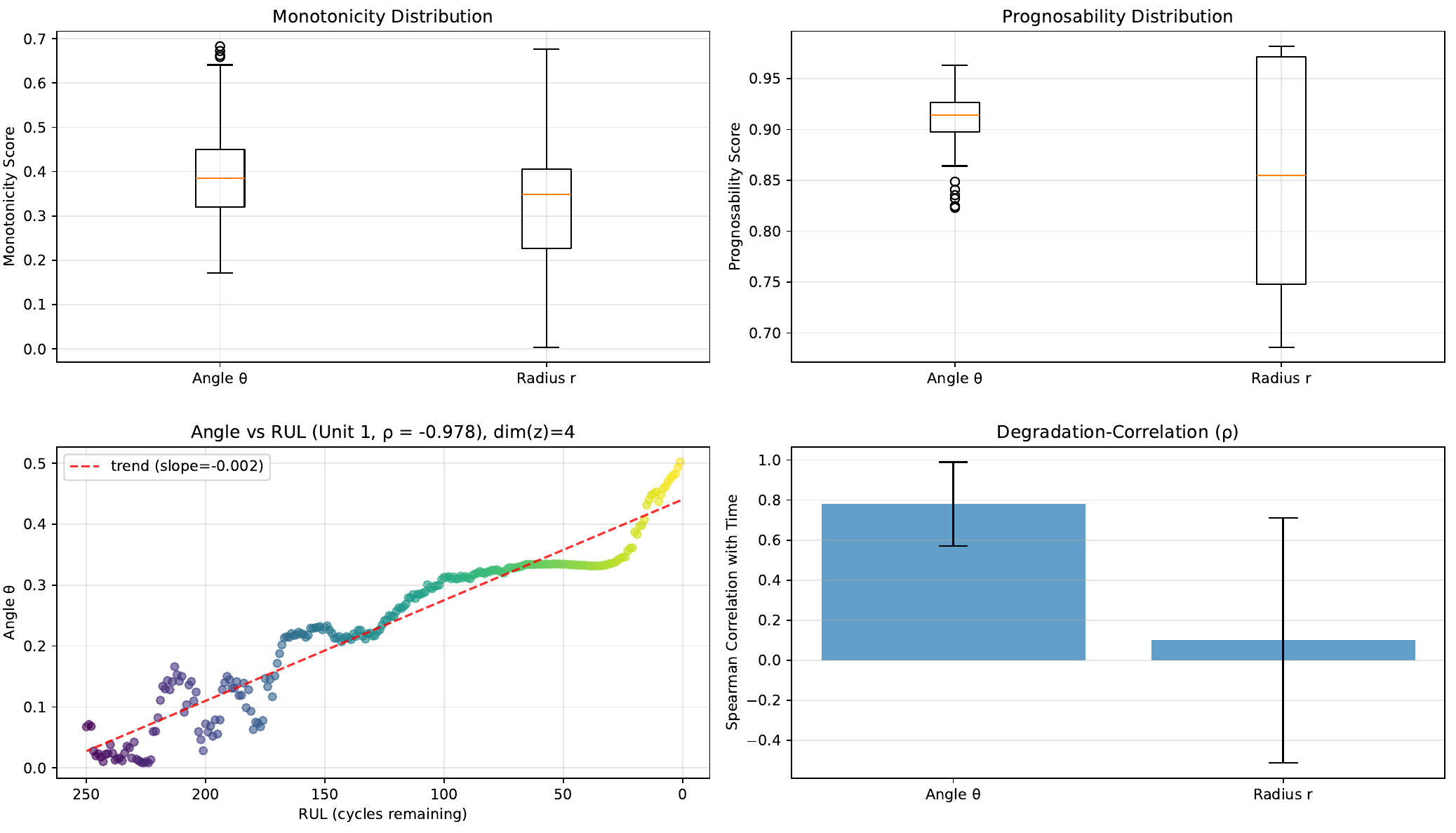}
        \caption{FD003, $\dim(z){=}4$.}
    \end{subfigure}
    \hfill
    \begin{subfigure}[b]{0.49\textwidth}
        \includegraphics[width=\linewidth]{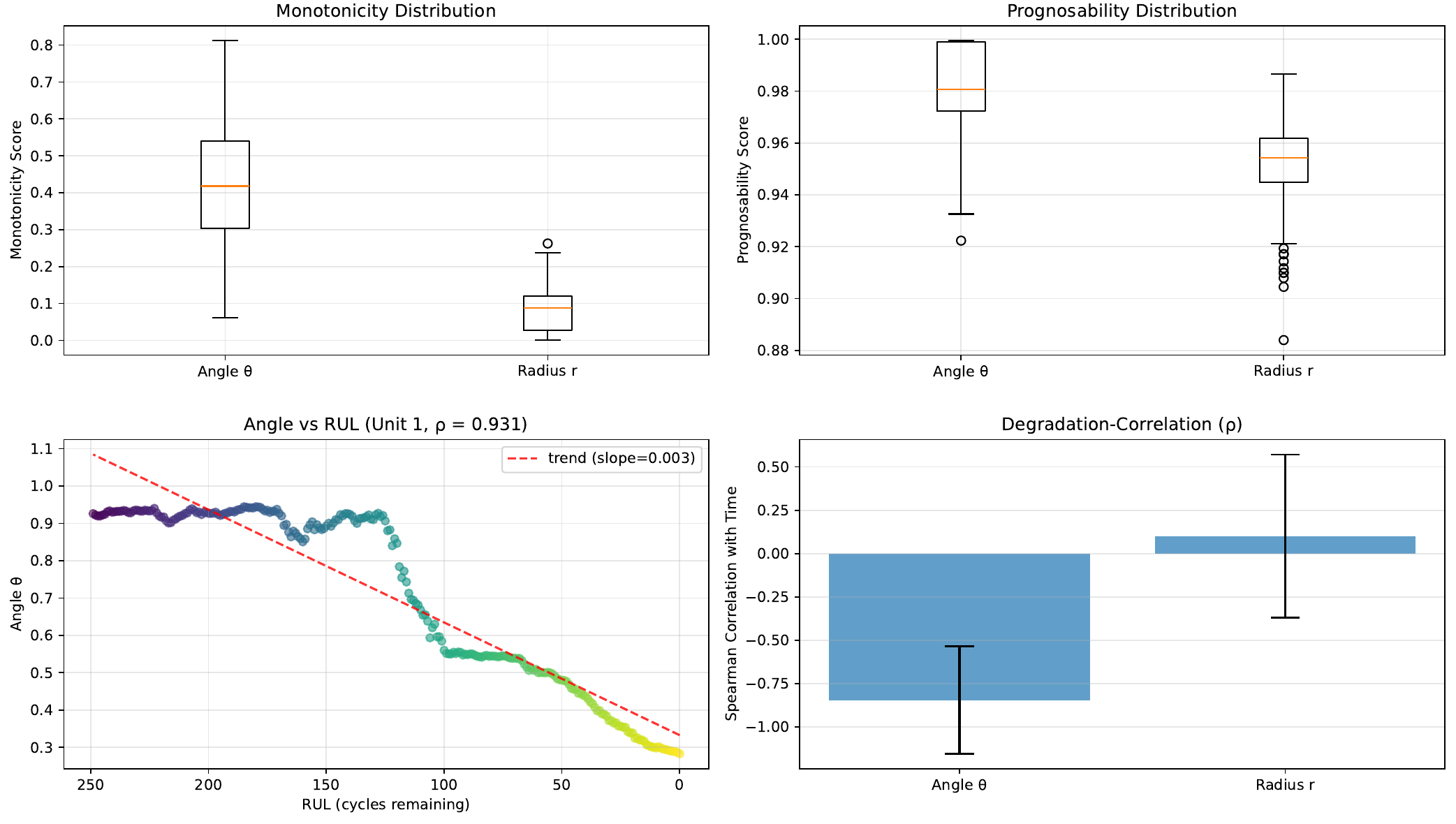}
        \caption{FD003, $\dim(z){=}16$.}
    \end{subfigure}\\[0.5em]
    \begin{subfigure}[b]{0.49\textwidth}
        \includegraphics[width=\linewidth]{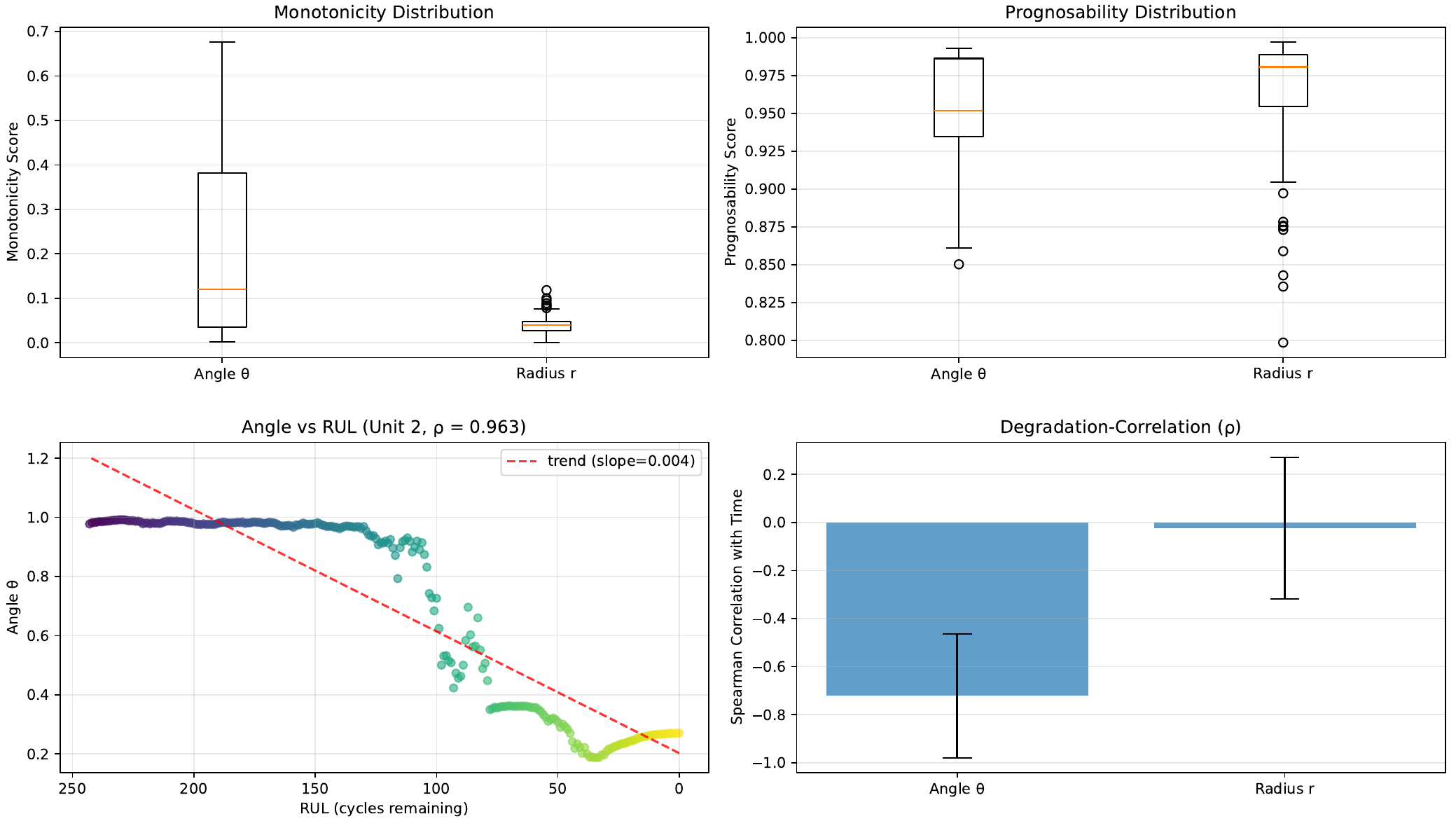}
        \caption{FD003, $\dim(z){=}32$.}
    \end{subfigure}
    \hfill
    \begin{subfigure}[b]{0.49\textwidth}
        \includegraphics[width=\linewidth]{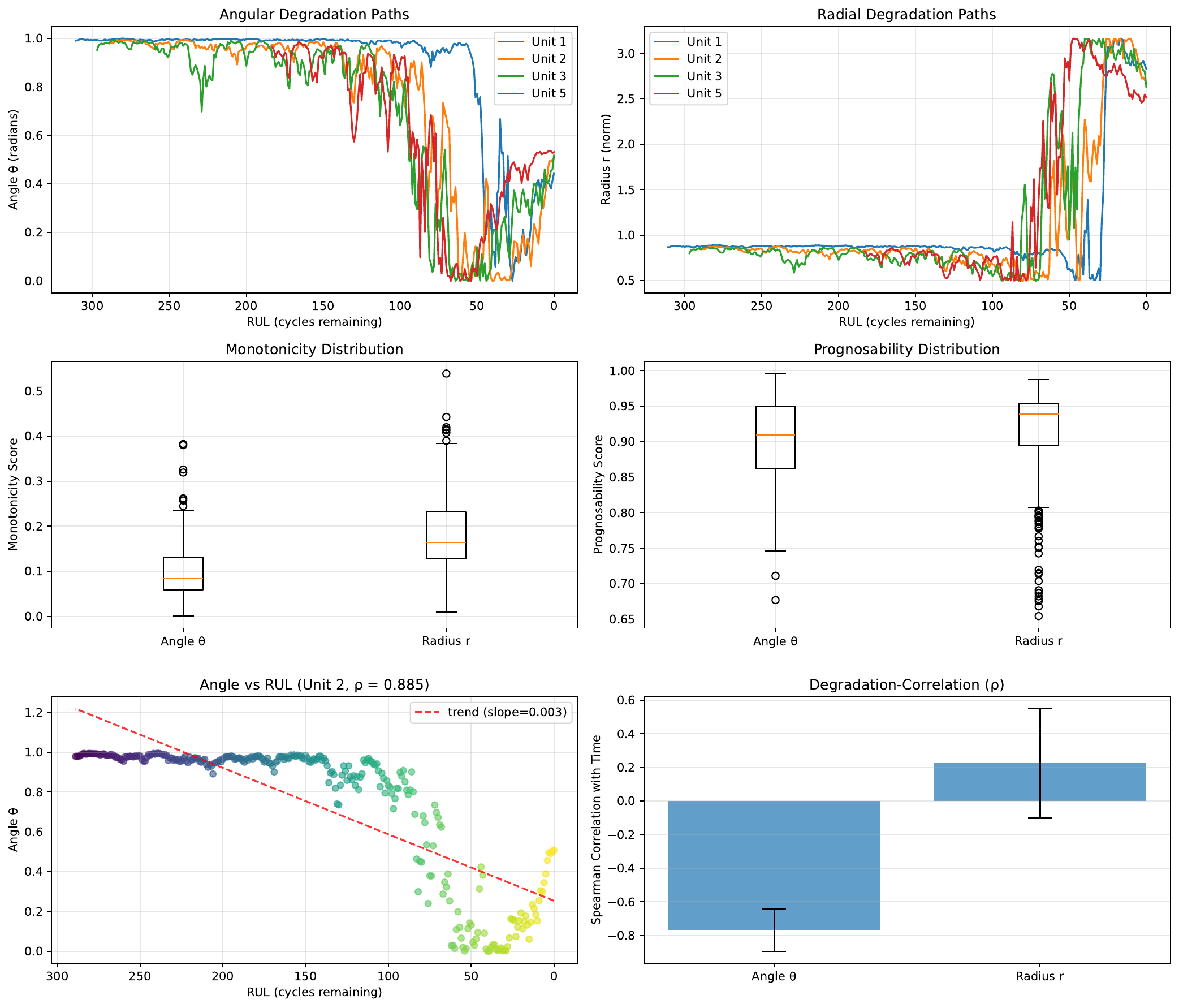}
        \caption{FD004, $\dim(z){=}32$.}
    \end{subfigure}
    \caption{\textbf{PHM diagnostics across latent dimensions}: angular and radial degradation paths, monotonicity and prognosability, angle versus RUL, and Spearman correlation.}
\end{figure}

\section{Mouse Neural Activity (Qualitative)}\label{appx:mouse}
The submitted version included a forecasting experiment on this single-trajectory recording, following the forward-simulation protocol of latent dynamics baselines~\citep{hu2025sing}. That protocol lets the encoder see the windows it later rolls out; under a strict temporal split (encoder and normalization fit on the first 70\%, rollout on the last 30\%), the 15-step $R^2$ drops from 0.32 to $-0.19$ (5 seeds), below a persistence baseline (0.90). We therefore removed the forecasting experiment. We keep only the qualitative latent topology below, for which the encoder is trained on the whole recording and no out-of-sample claim is made: STEP shows phase changes aligned with intruder entry and exit, as does SoftCLT, with a less ordered manifold.

\begin{figure}[h]
    \centering
    \begin{subfigure}{0.35\linewidth}
        \includegraphics[width=\linewidth]{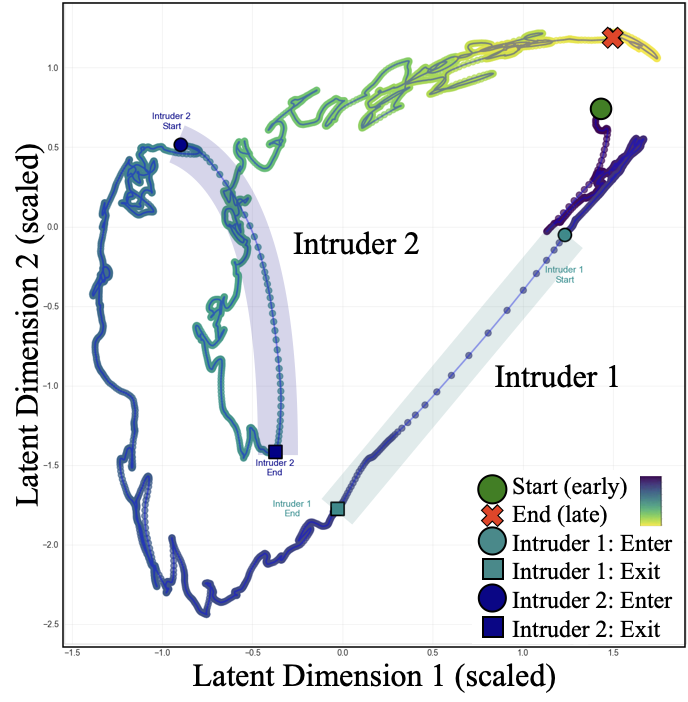}
        \caption{SoftCLT}
    \end{subfigure}\qquad
    \begin{subfigure}{0.35\linewidth}
        \includegraphics[width=\linewidth]{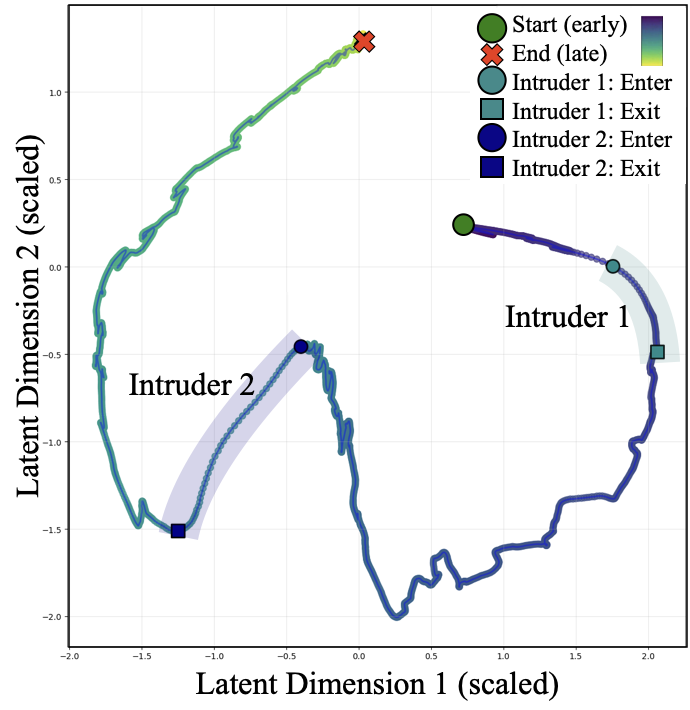}
        \caption{STEP}
    \end{subfigure}
    \caption{\textbf{Mouse brain activity}~\citep{Vinograd_Nair_Kim_Linderman_Anderson_2024}, $\bar{z}$ (t-SNE), qualitative.}
    \label{fig:vinograd_topology}
\end{figure}

\newpage
\section*{NeurIPS Paper Checklist}


\begin{enumerate}

\item {\bf Claims}
    \item[] Question: Do the main claims made in the abstract and introduction accurately reflect the paper's contributions and scope?
    \item[] Answer: \answerYes{}
    \item[] Justification: The abstract and introduction state the scope (progressive time series), the method, and the evidence: HI trendability on three degradation benchmarks, held-out controls on elapsed time, five-seed RUL results, and qualitative robotics results. Interpretability claims are limited to structured, predictive, linearly accessible coordinates (Sec.~1).


\item {\bf Limitations}
    \item[] Question: Does the paper discuss the limitations of the work performed by the authors?
    \item[] Answer: \answerYes{}
    \item[] Justification: Sec.~6 lists the progressive assumption, the need for complete run-to-failure training trajectories (sampler and prototypes), ambiguous or multiple end states, the limited meaning of the coordinates (radius not a mode label, 2D view dataset dependent), and the scope of evidence (no SOTA on FD001/FD003/FD004, qualitative robotics, no forecasting).


\item {\bf Theory assumptions and proofs}
    \item[] Question: For each theoretical result, does the paper provide the full set of assumptions and a complete (and correct) proof?
    \item[] Answer: \answerNA{}
    \item[] Justification: The paper makes no formal theoretical claims; the Markov formulation in Appendix~A is motivational.


    \item {\bf Experimental result reproducibility}
    \item[] Question: Does the paper fully disclose all the information needed to reproduce the main experimental results of the paper to the extent that it affects the main claims and/or conclusions of the paper (regardless of whether the code and data are provided or not)?
    \item[] Answer: \answerYes{}
    \item[] Justification: Architecture, losses, sampler, hyperparameters, protocols and baselines are given in Secs.~3--4 and Appendices~B--C; code is released.


\item {\bf Open access to data and code}
    \item[] Question: Does the paper provide open access to the data and code, with sufficient instructions to faithfully reproduce the main experimental results, as described in supplemental material?
    \item[] Answer: \answerYes{}
    \item[] Justification: Code is available at https://github.com/LucasStill/STEP with scripts to reproduce the experiments. All datasets are public (C-MAPSS, N-CMAPSS, NASA battery, GR00T, LeRobot, and the mouse recording of Vinograd et al.).


\item {\bf Experimental setting/details}
    \item[] Question: Does the paper specify all the training and test details (e.g., data splits, hyperparameters, how they were chosen, type of optimizer) necessary to understand the results?
    \item[] Answer: \answerYes{}
    \item[] Justification: Splits follow the standard C-MAPSS protocol and the CruiseBench protocol for N-CMAPSS; battery cell selection, hyperparameters, and head training are described in Sec.~4 and Appendices~B--C.


\item {\bf Experiment statistical significance}
    \item[] Question: Does the paper report error bars suitably and correctly defined or other appropriate information about the statistical significance of the experiments?
    \item[] Answer: \answerYes{}
    \item[] Justification: Main results are mean $\pm$ std over 5 seeds (Tables~1--3); unit-level bootstrap 95\% confidence intervals are reported for HI comparisons and in Appendix~D. Single-run results (prototype ablation, noise robustness, feature ablation) are labelled as such.


\item {\bf Experiments compute resources}
    \item[] Question: For each experiment, does the paper provide sufficient information on the computer resources (type of compute workers, memory, time of execution) needed to reproduce the experiments?
    \item[] Answer: \answerYes{}
    \item[] Justification: Appendix~B reports the hardware and time per run. Models are small (latent dimension 2--32, $d_\text{model}=32$); each run fits on a single GPU.

    
\item {\bf Code of ethics}
    \item[] Question: Does the research conducted in the paper conform, in every respect, with the NeurIPS Code of Ethics \url{https://neurips.cc/public/EthicsGuidelines}?
    \item[] Answer: \answerYes{} 
    \item[] Justification: Yes it conforms with the code of ethics. 

\item {\bf Broader impacts}
    \item[] Question: Does the paper discuss both potential positive societal impacts and negative societal impacts of the work performed?
    \item[] Answer: \answerYes{}
    \item[] Justification: Better health indicators can support predictive maintenance in safety-critical systems. A risk is over-trust: the learned coordinates are predictive features, not validated physical health measures, and Sec.~6 states that any safety-critical use requires domain validation.

    
\item {\bf Safeguards}
    \item[] Question: Does the paper describe safeguards that have been put in place for responsible release of data or models that have a high risk for misuse (e.g., pre-trained language models, image generators, or scraped datasets)?
    \item[] Answer: \answerNA{} 
    \item[] Justification: All datasets are open-sourced and freely accessible, the architectures and methods do not create any need for safeguards. 

\item {\bf Licenses for existing assets}
    \item[] Question: Are the creators or original owners of assets (e.g., code, data, models), used in the paper, properly credited and are the license and terms of use explicitly mentioned and properly respected?
    \item[] Answer: \answerYes{} 
    \item[] Justification: All datasets and models used are properly credited to their authors, and results accurately reported. 

\item {\bf New assets}
    \item[] Question: Are new assets introduced in the paper well documented and is the documentation provided alongside the assets?
    \item[] Answer: \answerYes{}
    \item[] Justification: We release the code with documentation under the licence stated in the repository.


\item {\bf Crowdsourcing and research with human subjects}
    \item[] Question: For crowdsourcing experiments and research with human subjects, does the paper include the full text of instructions given to participants and screenshots, if applicable, as well as details about compensation (if any)? 
    \item[] Answer: \answerNA{} 
    \item[] Justification: No crowdsourcing nor human subjects were involved in this work. 

\item {\bf Institutional review board (IRB) approvals or equivalent for research with human subjects}
    \item[] Question: Does the paper describe potential risks incurred by study participants, whether such risks were disclosed to the subjects, and whether Institutional Review Board (IRB) approvals (or an equivalent approval/review based on the requirements of your country or institution) were obtained?
    \item[] Answer: \answerNA{} 
    \item[] Justification: No crowdsourcing nor human subjects were involved in this work. 

\item {\bf Declaration of LLM usage}
    \item[] Question: Does the paper describe the usage of LLMs if it is an important, original, or non-standard component of the core methods in this research? Note that if the LLM is used only for writing, editing, or formatting purposes and does \emph{not} impact the core methodology, scientific rigor, or originality of the research, declaration is not required.
    \item[] Answer: \answerNA{} 
    \item[] Justification: Nothing to declare, LLMs are not part of this work.

\end{enumerate}

\end{document}